\documentclass[10pt,twocolumn,letterpaper]{article}

\usepackage{iccv}              
\usepackage{multicol, multirow, booktabs}
\usepackage{pifont}

\usepackage{bbm}
\usepackage{array}
\usepackage{subcaption}
\usepackage{graphicx}
\usepackage{pgfplots}
\usepackage{tikz}
\usetikzlibrary{patterns}
\usepackage{pgfplots}
\usetikzlibrary{patterns, pgfplots.groupplots}
\usepgfplotslibrary{groupplots}
\usepackage{pgfplotstable}
\usepackage{adjustbox}
\usepackage{overpic}
\usetikzlibrary{positioning, arrows}
\usepackage{wrapfig}
\usepackage{tcolorbox} 
\usepackage{forest}
\usepackage{fontawesome5}
\usepackage{amsmath}
\usetikzlibrary{decorations.pathreplacing}
\usetikzlibrary{positioning, fit, arrows.meta, shapes.multipart, shapes.geometric}

\definecolor{iccvblue}{rgb}{0.21,0.49,0.74}

\usepackage[pagebackref,breaklinks,colorlinks,allcolors=iccvblue]{hyperref}
\usepackage{amsmath}
\usepackage{tikz}
\usepackage{dsfont}
\usepackage{amsmath, amssymb}

\def\paperID{5585} 
\def\confName{ICCV}
\def\confYear{2025}

\title{ArgoTweak: Towards Self-Updating HD Maps through Structured Priors}

\author{
Lena Wild$^{1,2}$ \qquad 
Rafael Valencia$^{2}$ \qquad 
Patric Jensfelt$^{1}$\\
{$^1$KTH Royal Institute of Technology}
\quad
$^2$TRATON \ \\
{\tt\small \{lwild, patric\}@kth.se} \quad {\tt\small rafael.valencia.carreno@se.traton.com}}

\begin{document}  
\maketitle
\begin{abstract}
Reliable integration of prior information is crucial for self-verifying and self-updating HD maps. However, no public dataset includes the required triplet of prior maps, current maps, and sensor data. As a result, existing methods must rely on synthetic priors, which create inconsistencies and lead to a significant sim2real gap.
To address this, we introduce ArgoTweak, the first dataset to complete the triplet with realistic map priors. At its core, ArgoTweak employs a bijective mapping framework, breaking down large-scale modifications into fine-grained atomic changes at the map element level, thus ensuring interpretability. This paradigm shift enables accurate change detection and integration while preserving unchanged elements with high fidelity.
Experiments show that training models on ArgoTweak significantly reduces the sim2real gap compared to synthetic priors. Extensive ablations further highlight the impact of structured priors and detailed change annotations. By establishing a benchmark for explainable, prior-aided HD mapping, ArgoTweak advances scalable, self-improving mapping solutions. The dataset, baselines, map modification toolbox, and further resources are available at \href{https://kth-rpl.github.io/ArgoTweak/}{https://KTH-RPL.github.io/ArgoTweak/}.
\end{abstract}    
\section{Introduction}
\label{sec:intro}

High-definition (HD) maps are essential for autonomous driving, offering precise lane-level information for long-horizon predictions and occlusions handling \cite{hdmaps}. Traditionally, these HD maps have been created through offline manual annotation -- a process that, while accurate, is both labor-intensive and geographically constrained. This inherent lack of scalability has spurred a shift toward automated, end-to-end HD map generation, where bird’s-eye-view (BEV) feature backbones predict map structures directly from sensor data \cite{HDMapNet, vectormapnet, MapTR}. 

\begin{figure}[t!]
    \centering
    \hspace{-0.2cm}
    \begin{tikzpicture}[
        node distance=0.5cm and 0.2cm, 
        every node/.style={font=\footnotesize,rounded corners, line width=0.005cm  },
        ]
    \definecolor{ribbonColor}{RGB}{230, 230, 230}
    \definecolor{mpl_green}{RGB}{0,128,0}
    \definecolor{mpl_green_}{RGB}{255, 165, 0}
    \definecolor{mpl_blue}{RGB}{0, 0, 255}
    \definecolor{mplred}{rgb}{1,0,0}
    \definecolor{mplblue}{rgb}{0.5, 0.5, 0.5}
     \node[minimum width=2cm,yshift=-0.5cm, xshift=-2cm] (sensor) 
{\includegraphics[width=1.5cm]{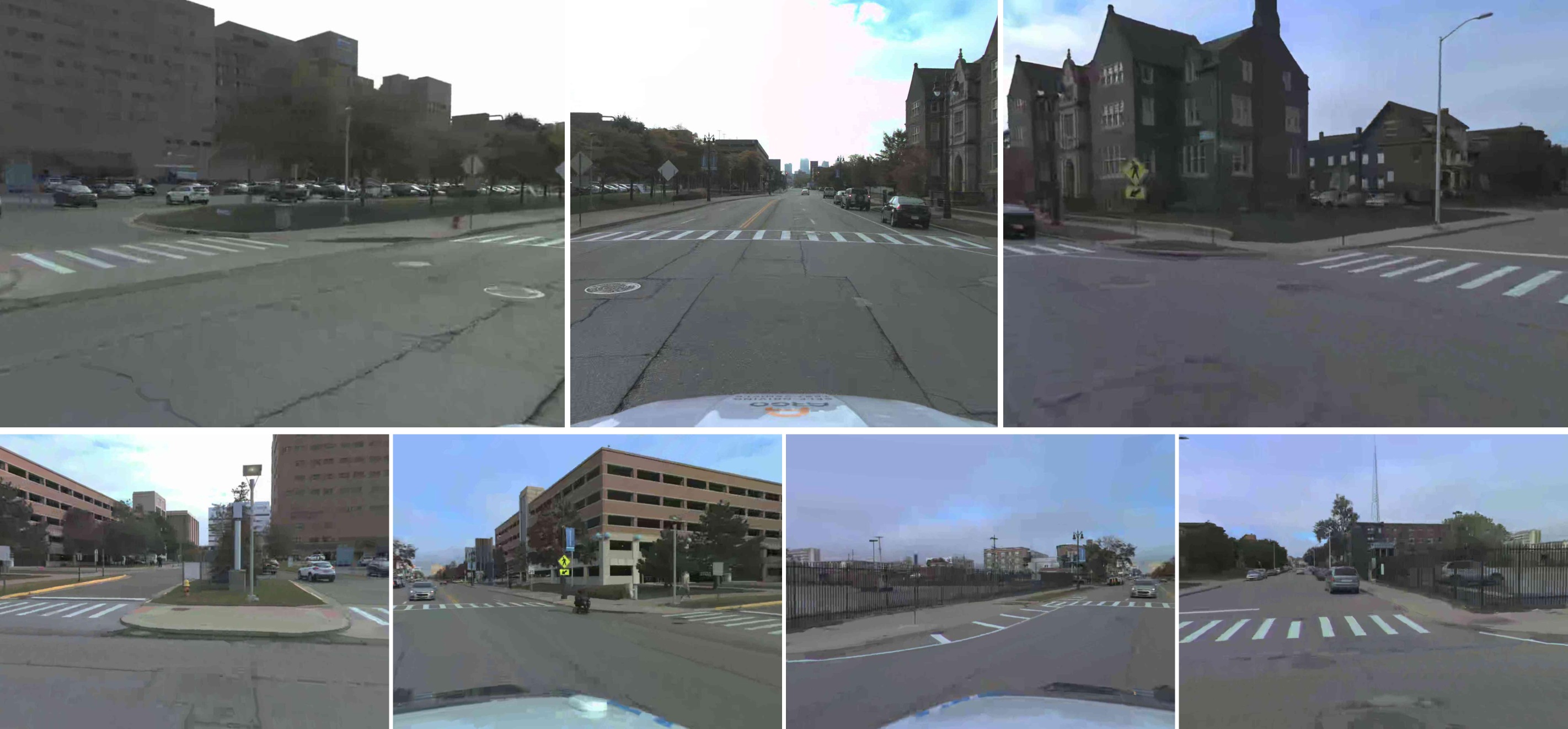}};
        \node[minimum width=2cm,above=-0.2cm of sensor.north]
{sensor input};

\node[minimum width=3cm, minimum height=1cm, fill=mpl_green!30, draw=mpl_green,right=0.3cm of sensor,yshift=-0.4cm] (network)
    {\begin{tabular}{c} \text{Explainable} \\ \footnotesize \text{Prior-Aided Mapping} \end{tabular}};
    \node[minimum width=8cm, minimum height=0.7cm,fill=mplblue!20,below=1.5cm of network, yshift=-0.4cm, xshift=0.325cm] (argotweak){\hspace{-0.3cm}\includegraphics[width=0.5cm]{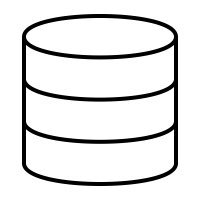} \raisebox{0.1cm}{\textbf{ArgoTweak}}}; 
\node (network_east) at ([xshift=1cm]network.west) {};
          \begin{scope}[shift={(network.west)},xshift=-3.7cm,  yshift=-3.8cm, line width=0.5pt, transform canvas={scale=0.38}]
           \draw[ draw=white, draw=gray]  (-1.5,-2.2) rectangle (1.5,2.2);
\node[scale = 1/0.38] at (0,-2.7) {\footnotesize map prior}; 

            \fill[ribbonColor] 
                (-1,-2) to[out=90,in=-70] (-1.2,2)
                -- (-0.2,2) to[out=-70,in=90] (-0,-2) -- cycle;
            \draw[thick,dashed, mpl_green] 
                (-0,-2) to[out=90,in=-70] (-0.2,2);
            \draw[thick, mpl_green] 
                (-1,-2) to[out=90,in=-70] (-1.2,2);
            \draw[-<, thick, mpl_green_,line width=1pt, scale=1] 
                (-0.7,1.8) to[out=90,in=-80] (-0.5,1); 
                (-0.5,-2) to[out=90,in=-90] (-0.5,2); 
                (-0,-2) to[out=90,in=-90] (-0,2); 
                (-0.25,1.8) to[out=90,in=-90] (-0.25,1);

            \fill[ribbonColor] 
                (1,0) to[out=90,in=-70] (0.8,2)
                -- (0.3,2) to[out=-70,in=90] (0.5,0) -- cycle;
            \draw[thick,dashed,  mpl_green] 
                (0.5,0) to[out=90,in=-70] (0.3,2);
            \draw[thick, mpl_green] 
                (1,0) to[out=90,in=-70] (0.8,2);
            \draw[>-, thick, mpl_green_] 
                (0.75,1) to[out=90,in=-90] (0.75,-0.8);
                
            \fill[ribbonColor] 
                (-0,0) to[out=90,in=-70] (-0.2,2)
                -- (0.3,2) to[out=-70,in=90] (0.5,0) -- cycle;
            \draw[thick,dashed, mpl_green] 
                (0.5,0) to[out=90,in=-70] (0.3,2);
            \draw[thick, mpl_green] 
                (-0,0) to[out=90,in=-70] (-0.2,2);
            \draw[>-, thick, mpl_green_] 
                (0.25,1) to[out=90,in=-90] (0.25,-0.8);

            \fill[ribbonColor] 
                (1,-2) to[out=90,in=90] (1,0)
                -- (0.5,0) to[out=90,in=90] (0.5,-2) -- cycle;
            \draw[thick,solid,  mpl_green] 
                (0.5,-2) to[out=90,in=-90] (0.5,0);
            \draw[thick, mpl_green] 
                (1,-2) to[out=90,in=-90] (1,0);
            \draw[>-, thick, mpl_green_] 
                (0.75,-1) to[out=90,in=-90] (0.75,-1.8);
        
            \fill[ribbonColor] 
                (-0,-2) to[out=90,in=90] (-0,0)
                -- (0.5,0) to[out=90,in=90] (0.5,-2) -- cycle;
            \draw[thick,solid, mpl_green] 
                (0.5,-2) to[out=90,in=-90] (0.5,0);
            \draw[thick, mpl_green] 
                (-0,-2) to[out=90,in=-90] (-0,0);
            \draw[>-, thick, mpl_green_] 
                (0.25,-1) to[out=90,in=-90] (0.25,-1.8);
            
          \end{scope}
          \begin{scope}[shift={(network.west)},xshift=9.4cm, yshift=-1.5cm,line width=0.5pt, transform canvas={scale=0.38}]
    
            \definecolor{ribbonColor}{RGB}{230, 230, 230}
    \definecolor{mpl_green}{RGB}{0,128,0}
    \definecolor{mpl_green_}{RGB}{255, 165, 0}
    \definecolor{mpl_blue}{RGB}{0, 0, 255}
           \draw[ fill=white,draw=gray]  (-1.5,-2.2) rectangle (1.5,2.2);
           \draw[ fill=gray!10, rounded corners]  (-1.5,-3) rectangle (4.7,-2.4);
\fill[green] (-1.3,-2.8) rectangle (-1.0,-2.6);
\node[right,scale = 1/0.8] at (-0.9,-2.7) {Insertion};

\fill[mpl_green] (0.7,-2.8) rectangle (1.0,-2.6);
\node[right,scale = 1/0.8] at (1.1,-2.7) {Geometry};

\fill[violet] (2.8,-2.8) rectangle (3.1,-2.6);
\node[right,scale = 1/0.8] at (3.15,-2.7) {Marking};
           
\node[scale = 1/0.38] at (1.5,2.7) {\footnotesize explainable update}; 
            \fill[ribbonColor] 
                (-1,-2) to[out=90,in=-90] (-1,2)
                -- (-0.5,2) to[out=-90,in=90] (-0.5,-2) -- cycle;
            \draw[thick,dashed, mpl_green] 
                (-0,-2) to[out=90,in=-90] (-0,2);
            \draw[thick, mpl_green] 
                (-1,-2) to[out=90,in=-90] (-1,2);
            \draw[-<, thick, mpl_green_,line width=1pt, scale=1] 
                (-0.75,1.8) to[out=90,in=-90] (-0.75,1);

            \fill[ribbonColor] 
             (-0,-2) to[out=90,in=-90] (-0,2)
             -- (-0.5,2) to[out=-90,in=90] (-0.5,-2) -- cycle;
            \draw[thick, dashed,mpl_green] 
                (-0.5,-2) to[out=90,in=-90] (-0.5,2);
            \draw[thick, mpl_green] 
                (-0,-2) to[out=90,in=-90] (-0,2);
            \draw[-<, thick, mpl_green_] 
                (-0.25,1.8) to[out=90,in=-90] (-0.25,1);

            \fill[ribbonColor] 
                (1,0) to[out=90,in=-90] (1,2)
                -- (0.5,2) to[out=-90,in=90] (0.5,0) -- cycle;
            \draw[thick,dashed,  mpl_green] 
                (0.5,0) to[out=90,in=-90] (0.5,2);
            \draw[thick, mpl_green] 
                (1,0) to[out=90,in=-90] (1,2);
            \draw[>-, thick, mpl_green_] 
                (0.75,1) to[out=90,in=-90] (0.75,-0.8);
                
            \fill[ribbonColor] 
                (-0,0) to[out=90,in=-90] (-0,2)
                -- (0.5,2) to[out=-90,in=90] (0.5,0) -- cycle;
            \draw[thick,dashed, mpl_green] 
                (0.5,0) to[out=90,in=-90] (0.5,2);
            \draw[thick, mpl_green] 
                (-0,0) to[out=90,in=-90] (-0,2);
            \draw[>-, thick, mpl_green_] 
                (0.25,1) to[out=90,in=-90] (0.25,-0.8);

            \fill[ribbonColor] 
                (1,-2) to[out=90,in=90] (1,0)
                -- (0.5,0) to[out=90,in=90] (0.5,-2) -- cycle;
            \draw[thick,dashed,  mpl_green] 
                (0.5,-2) to[out=90,in=-90] (0.5,0);
            \draw[thick, mpl_green] 
                (1,-2) to[out=90,in=-90] (1,0);
            \draw[>-, thick, mpl_green_] 
                (0.75,-1) to[out=90,in=-90] (0.75,-1.8);
        
            \fill[ribbonColor] 
                (-0,-2) to[out=90,in=90] (-0,0)
                -- (0.5,0) to[out=90,in=90] (0.5,-2) -- cycle;
            \draw[thick,dashed, mpl_green] 
                (0.5,-2) to[out=90,in=-90] (0.5,0);
            \draw[thick, mpl_green] 
                (-0,-2) to[out=90,in=-90] (-0,0);
            \draw[>-, thick, mpl_green_] 
                (0.25,-1) to[out=90,in=-90] (0.25,-1.8);
           
                 \fill[white,opacity=0.9,  draw=mplblue] 
                (1.3,0.) to[] (-1.3,0.05)
                -- (-1.3,0.7) to[] (1.3,0.6) -- cycle;
          
          \end{scope}
          
          \begin{scope}[shift={(network.west)},xshift=12.7cm, yshift=-1.5cm,line width=0.5pt, transform canvas={scale=0.38}]
           
            \definecolor{ribbonColor}{RGB}{230, 230, 230}
    \definecolor{mpl_green}{RGB}{0,128,0}
    \definecolor{mpl_green_}{RGB}{255, 165, 0}
    \definecolor{mpl_blue}{RGB}{0, 0, 255}
           \draw[ draw=white, draw=gray]  (-1.5,-2.2) rectangle (1.5,2.2);
   
\node[scale = 1/0.38] at (0,2.7) {\footnotesize  }; 
            \fill[green, opacity=0.3] 
                (-1,-2) to[out=90,in=-90] (-1,2)
                -- (-0.5,2) to[out=-90,in=90] (-0.5,-2) -- cycle;
            \draw[thick, violet] 
                (-0,-2) to[out=90,in=-90] (-0,2);
            \draw[thick, gray] 
                (-1,-2) to[out=90,in=-90] (-1,2); 
                (-0.75,1.8) to[out=90,in=-90] (-0.75,1);
                
            \fill[violet!30] 
             (-0,-2) to[out=90,in=-90] (-0,2)
             -- (-0.5,2) to[out=-90,in=90] (-0.5,-2) -- cycle;
            \draw[thick,violet] 
                (-0.5,-2) to[out=90,in=-90] (-0.5,2);
            \draw[thick, gray] 
                (-0,-2) to[out=90,in=-90] (-0,2); 
                (-0.25,1.8) to[out=90,in=-90] (-0.25,1);
    
            \fill[ribbonColor] 
                (1,0) to[out=90,in=-90] (1,2)
                -- (0.5,2) to[out=-90,in=90] (0.5,0) -- cycle;
            \draw[thick,  gray] 
                (0.5,0) to[out=90,in=-90] (0.5,2);
            \draw[thick, gray] 
                (1,0) to[out=90,in=-90] (1,2); 
                (0.75,1) to[out=90,in=-90] (0.75,-0.8);
                
            \fill[ribbonColor] 
                (-0,0) to[out=90,in=-90] (-0,2)
                -- (0.5,2) to[out=-90,in=90] (0.5,0) -- cycle;
            \draw[thick, gray] 
                (0.5,0) to[out=90,in=-90] (0.5,2);
            \draw[thick, gray] 
                (-0,0) to[out=90,in=-90] (-0,2); 
                (0.25,1) to[out=90,in=-90] (0.25,-0.8);

            \fill[violet!30] 
                (1,-2) to[out=90,in=90] (1,0)
                -- (0.5,0) to[out=90,in=90] (0.5,-2) -- cycle;
            \draw[thick,  violet] 
                (0.5,-2) to[out=90,in=-90] (0.5,0);
            \draw[thick, gray] 
                (1,-2) to[out=90,in=-90] (1,0); 
                (0.75,-1) to[out=90,in=-90] (0.75,-1.8);
        
            \fill[violet!30] 
                (-0,-2) to[out=90,in=90] (-0,0)
                -- (0.5,0) to[out=90,in=90] (0.5,-2) -- cycle;
            \draw[thick, violet] 
                (0.5,-2) to[out=90,in=-90] (0.5,0);
            \draw[thick, gray] 
                (-0,-2) to[out=90,in=-90] (-0,0); 
                (0.25,-1) to[out=90,in=-90] (0.25,-1.8);

            \fill[mpl_green, opacity=0.4] 
                (1,0) to[out=90,in=-70] (0.8,2)
                -- (0.3,2) to[out=-70,in=90] (0.5,0) -- cycle;
            \draw[thick,dotted,  mpl_green] 
                (0.5,0) to[out=90,in=-70] (0.3,2);
            \draw[thick,dotted, mpl_green] 
                (1,0) to[out=90,in=-70] (0.8,2); 
                (0.75,1) to[out=90,in=-90] (0.75,-0.8);
                
            \fill[mpl_green, opacity=0.4] 
                (-0,0) to[out=90,in=-70] (-0.2,2)
                -- (0.3,2) to[out=-70,in=90] (0.5,0) -- cycle;
            \draw[thick,dotted, mpl_green] 
                (0.5,0) to[out=90,in=-70] (0.3,2);
            \draw[thick,dotted, mpl_green] 
                (-0,0) to[out=90,in=-70] (-0.2,2); 
                (0.25,1) to[out=90,in=-90] (0.25,-0.8);
                
            \fill[mpl_green, opacity=0.4] 
                (-1,0) to[out=90,in=-70] (-1.2,2)
                -- (-0.2,2) to[out=-70,in=90] (-0,0) -- cycle;
            \draw[thick,dotted, mpl_green] 
                (-0,0) to[out=90,in=-70] (-0.2,2);
            \draw[thick, dotted, mpl_green] 
                (-1,0) to[out=90,in=-70] (-1.2,2);
            
                 \fill[green,opacity=0.3,  draw=mplblue] 
                (1.3,0.) to[] (-1.3,0.05)
                -- (-1.3,0.7) to[] (1.3,0.6) -- cycle;

          \end{scope}

    \draw[-{Triangle[width=3mm,length=5pt]}, draw=mplblue!40, line width=1mm, opacity=0.7] ([xshift=-2.8cm]argotweak.north) -- ([yshift=0.3cm,xshift=-2.8cm] argotweak.north);
    \draw[-{Triangle[width=3mm,length=5pt]}, draw=mplblue!40, line width=1mm, opacity=0.7] ([xshift=0.3cm]argotweak.north west) -- ([yshift=2.4cm,xshift=0.3cm] argotweak.north west);
    \draw[-{Triangle[width=3mm,length=5pt]}, draw=mplblue!40, line width=1mm, opacity=0.7] ([xshift=-0.3cm]argotweak.north east) -- ([yshift=1.2cm,xshift=-0.3cm] argotweak.north east);
    \draw[->, draw=mpl_green] (network.east) -- ([xshift=0.3cm] network.east);
    \draw[->, draw=mpl_green] ([xshift=-0.5cm, yshift=0.25cm]network.west) -- ( [yshift=0.25cm]network.west);
    \draw[->, draw=mpl_green] ([xshift=-0.35cm, yshift=-0.25cm]network.west) -- ( [yshift=-0.25cm]network.west);
    \draw[->, draw=mpl_green] ([xshift=2.15cm, yshift=-1.2cm]network.east) -- ([xshift=2.15cm,yshift=-1.9cm] network.east) --
    node[midway, xshift=0.8cm, yshift=0.35cm] {\includegraphics[width=1cm]{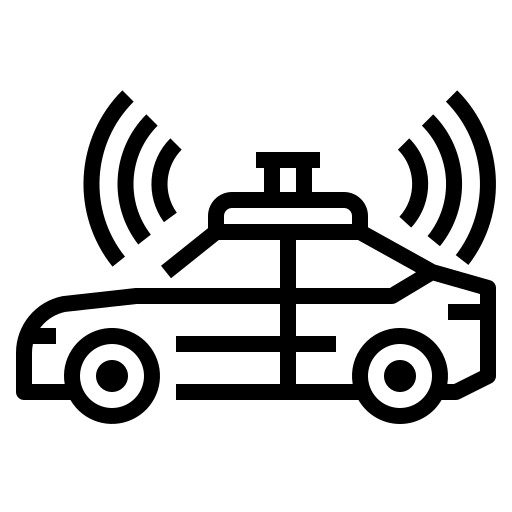}} 
    node[midway,  xshift=-1.5cm, yshift=0.275cm]{\includegraphics[width=1.2cm]{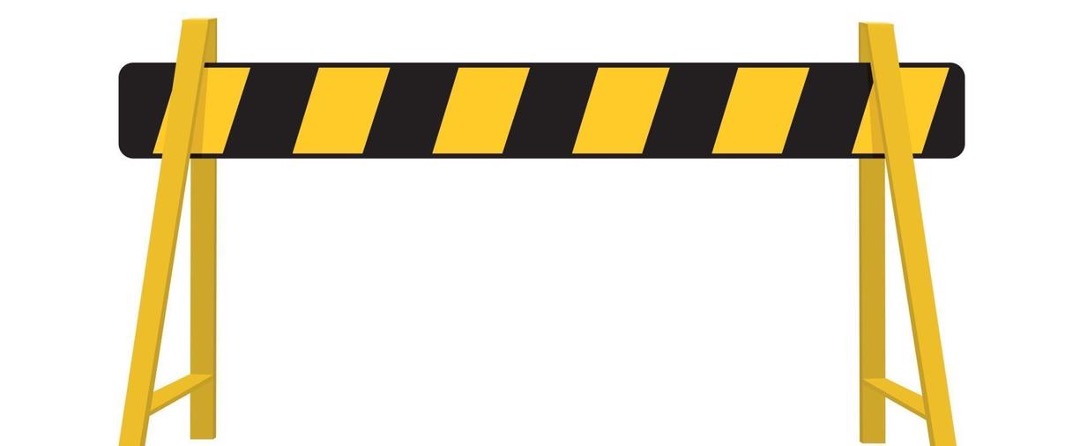}} 
    node[midway,  xshift=-1.6cm, yshift=0.9cm]{\includegraphics[width=0.6cm]{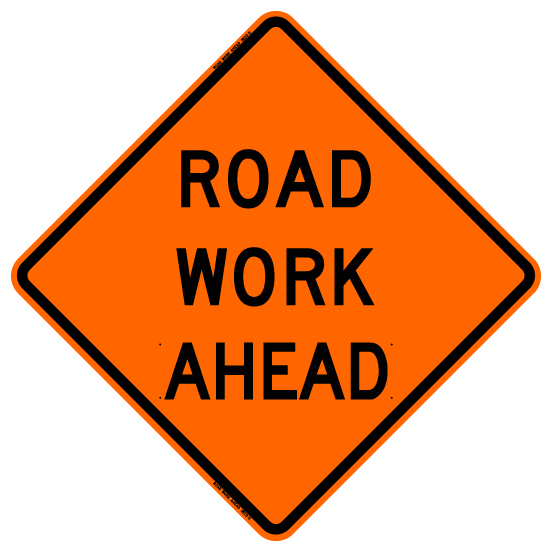}} 
    node[midway,  xshift=-1.6cm, yshift=0.25cm]{\includegraphics[width=0.7cm]{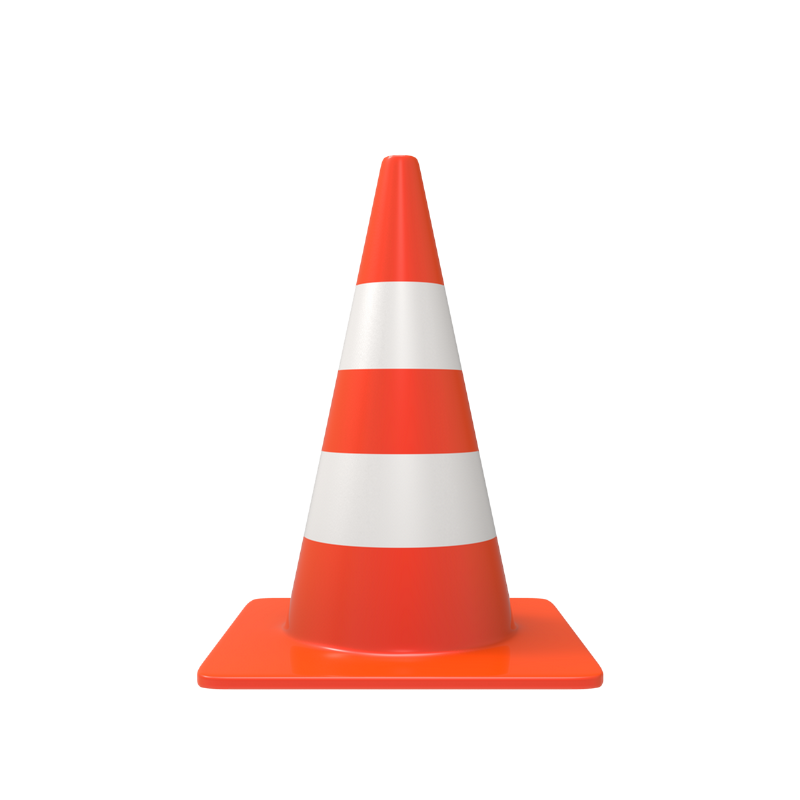} \hspace{-0.4cm} \includegraphics[width=0.7cm]{Img/main/Fig1/cone.png} \ \includegraphics[width=0.7cm]{Img/main/Fig1/cone.png}}([xshift=-0.3cm, yshift=-1.9cm]network.west);

    \draw[brown, line width=0.4mm] (0.17cm, -2.15cm) -- (0.17,-2.8cm);

    \end{tikzpicture}
    \caption{Overview of the ArgoTweak dataset and framework: Prior maps and current sensor input are integrated to enable self-updating HD mapping. Our annotations systematically decompose large-scale modifications into element-level changes, ensuring precise and explainable updates.}\label{fig:firstpage}
\end{figure}

While these automated methods reduce manual effort, challenges remain. Occlusions, sensor noise, and environmental variations impact reliability, and online-generated maps often lack the semantic richness and accuracy  of their offline counterparts \cite{bateman}. To bridge this gap, recent approaches integrate priors, such as outdated or less-accurate pre-existing maps, into the generation pipeline, as these lower-quality maps are typically readily available \cite{smerf, bateman, MtM, ExelMap, m3tr, PriorDrive}. Rather than constructing maps from scratch, these  models can refine inconsistencies in the priors, leading to promising improvements in map quality. 

Looking ahead, the role of map priors could extend beyond improving generation quality. In vehicles, HD maps are treated as a continuous source of information, but blindly trusting them is risky. Instead, autonomous vehicle fleets need to dynamically verify and update maps against online sensor data to ensure reliability. We argue that rather than treating map generation, change detection, and map updating as separate tasks, future approaches must unify these processes: By leveraging priors, enforcing consistency, and providing structured, explainable modifications at scale, HD maps could evolve into self-improving, continuously updating road representations.

Despite these exciting perspectives, research on integrating map priors remains constrained by data limitations. No public dataset currently provides the requisite triplet: a map prior, current sensor data, and an up-to-date ground-truth map. To compensate, existing approaches generate synthetic priors through scripted alterations, warping, or selective dropout of map elements \cite{bateman, m3tr, MtM, PriorDrive, ExelMap,tbv}.  This raises three key challenges: (1) The diversity of synthetic priors complicates method comparison, as each embeds different amounts of ground-truth information, with unclear effects on performance metrics.
(2) Current evaluation metrics fail to differentiate between performance on unchanged and newly updated map regions, making it unclear whether a model merely preserves existing structures or effectively updates the map. (3)	Synthetic perturbation methods do not capture the structured, semantically correlated nature of real-world changes. As studies suggest \cite{bateman, tbv}, this mismatch creates a significant simulation-to-reality (sim2real) gap, where models trained on synthetic priors struggle to generalize  to real-world scenarios.

To overcome these limitations and advance the vision of self-improving, continuously updating road representations, we introduce ArgoTweak -- the first hand-curated dataset featuring realistic map priors to align with up-to-date sensor data and ground-truth maps from the Argoverse 2 Map Change Dataset \cite{tbv} (\cref{fig:firstpage}). Additionally, we re-annotate real-world changes from \cite{tbv} to adhere to modern HD mapping standards, enabling the first evaluation of map prior integration approaches in real-world scenarios.

At the core of ArgoTweak is a bijective mapping framework that decomposes large-scale map modifications into atomic changes (\eg, insertions, deletions, and lane attribute updates) providing fine-grained, explainable labels that capture the nature and extent of modifications at the element level. This allows us to introduce a comprehensive metric that assesses the updated map in changed and unchanged regions separately, exposing the shortcomings of existing metrics in capturing map adaptation capabilities.

To demonstrate the value of our dataset, we train a baseline model, achieving a significantly reduced sim2real gap compared to existing approaches. Furthermore, we leverage the explainability of both our dataset and model to identify key challenges in integrating prior maps into modern HD mapping pipelines through detailed ablation studies. In summary, our main contributions are the following:
\begin{itemize}
    \item We introduce ArgoTweak, the first hand-curated dataset to complete the triplet of up-to-date sensor data and maps of \cite{tbv} with realistic map priors and refined real-world changes, enabling standardized training and evaluation of prior integration.
    \item We define a novel, comprehensive benchmark for update efficacy through our explainable, atomic change annotations, distinguishing pre-existing from updated regions.
    \item We propose a flexible baseline architecture for explainable prior-aided mapping and show a significantly reduced sim2real gap for ArgoTweak-trained models, while highlighting key challenges in prior integration through extensive ablation experiments.
\end{itemize}

\section{Related work}
\label{sec:relatedwork}
\subsection{Online mapping with and without prior}
HD map generation was initially framed as a semantic segmentation problem in the bird’s-eye view \cite{semantic_mapping}. However, the need for structured outputs led to vectorized approaches like HDMapNet \cite{HDMapNet}, which combined rasterized segmentation with vectorized representations. VectorMapNet \cite{vectormapnet} improved this with a two-stage pipeline modeling spatial relationships without heuristic post-processing. A major breakthrough came with MapTR \cite{MapTR} and MapTR-v2 \cite{maptrv2}, which reformulated map decoding as a one-stage DETR-like \cite{DETR} process, enhancing speed and accuracy. While MapTR-v2 remains a strong baseline, recent works advanced along three directions: (1) refining single-frame generation with improved queries and decoding \cite{InsMapper, Mask2Map, HiMap,ADMap,geomap, pointset,mgmap,mgmapnet}, (2) enhancing temporal consistency via multi-frame fusion and global map stitching \cite{MapTracker, StreamMapNet, SQNet,neuralmapprior, histomap, unveil}, and (3) enriching maps with topological and semantic details \cite{LaneSegNet, TopoNet, topomlp}.

All these methods rely solely on sensor data. However, as \cite{MtM} highlights, using outdated or lower-quality maps as a prior can improve results. Building on prior works incorporating standard definition maps \cite{LaneSegNet, smerf, PriorDrive, pmapnet}, \cite{MtM} and \cite{bateman} propose hybrid queries combining existing HD map elements with learnable ones. PriorDrive \cite{PriorDrive} employs hybrid prior embedding and dual encoding for SD and lower-quality HD maps, while M3TR \cite{m3tr} adapts to varying map priors. In our previous work, ExelMap \cite{ExelMap}, we introduced an explainable, element-based approach for identifying and updating changed map elements.

\subsection{Datasets and challenges}
The availability of datasets varies across research areas. For prior-less online mapping, public datasets such as nuScenes~\cite{nuscenes}, Argoverse 2~\cite{Argoverse2}, and OpenLane-v2~\cite{wang2023openlanev2} provide key resources, despite concerns over geographically overlapping splits \cite{lilja, StreamMapNet}. These datasets, both in their original and re-grouped forms, remain widely used (\cf~\cref{tab:dataset_specs}).

In contrast, prior-aided HD mapping lacks public datasets that include map priors, current sensor data, and up-to-date ground-truth maps. To address this, researchers synthesize priors through discrete modifications (e.g., element dropout, duplication) \cite{bateman, m3tr, MtM, PriorDrive}, continuous transformations (e.g., noise injection, warping) \cite{bateman, MtM}, or rule-based approaches mimicking the semantically correlated nature of real-world changes \cite{MtM, tbv, ExelMap}.

However, existing synthetic prior generation techniques exhibit major limitations. Since studies selectively modify map elements in arbitrary ways, results from performance evaluations are not directly comparable. Furthermore, common evaluation metrics like mean average precision fail to distinguish between preserving existing structures and correctly updating outdated regions, making it unclear whether models genuinely adapt to real-world changes or simply maintain known elements \cite{ExelMap, m3tr}. A third issue is related to the reliance on synthetic perturbations, which introduces a sim2real  gap where models trained on artificial modifications struggle to generalize to real-world updates \cite{bateman}.

\section{Motivation}\label{sec:motivation}
In real-world environments, changes are typically localized and relatively rare, yet their effects can be highly consequential for safety \cite{tbv}. Hence, although most of the map remains stable over time, even minor undetected modifications can lead to critical failures. A robust mapping approach must balance two competing objectives: accurately detecting and integrating changes, and preserving unchanged elements with high fidelity.

We argue that achieving this balance requires more than just the triplet of standardized priors, sensor data, and up-to-date ground truth -- it necessitates a paradigm shift encompassing data representation, training methodologies, and evaluation protocols. To reliably integrate priors into mapping, we hence introduce the \textbf{ArgoTweak dataset}, the first to provide explicitly annotated priors using a novel \textbf{bijective change mapping framework} (\cref{sec:dataset}). On the model side, we propose an \textbf{explainable prior-aided mapping network} (\cref{sec:method}) that makes changes interpretable. To ensure rigorous evaluation, we introduce a \textbf{fine-grained metric} that quantifies both stability in unchanged regions and responsiveness to updates (\cref{sec:metric}).
Finally, we integrate our dataset, metric, and model (\cref{sec:experiments}) to validate our claims and establish a \textbf{new benchmark for explainable prior-aided HD mapping}.

\section{The ArgoTweak Dataset}
\label{sec:dataset}
\subsection{Bijective change mapping}
As our first contribution, we introduce \textbf{bijective change mapping}, a novel framework to systematically provide change annotations for each map element. While not a strict mathematical bijection, it is a design principle that guides how we relate high-level \emph{structural updates} -- complex modifications to road geometry or semantics -- to fine-grained \emph{atomic changes} applied at the element level (i.e., individual lane segments or pedestrian crossings in the widely used Argoverse 2 HD map format \cite{Argoverse2}).

Our motivation is twofold. First, we want any local road change, no matter how complex, to be decomposable into a unique and traceable set of edits on individual elements, to facilitate map learning. Second, to reduce annotation ambiguity, road layout changes must be annotated in a consistent manner, avoiding arbitrary distinctions between full element replacement and incremental modification. Enforcing this consistency is key for interpretability and model supervision.
  
\subsubsection*{Bijectivity requirements}\label{sec:bijectivityrecs}  \ We formally define atomic changes as $a_i$ with set $\mathbf{A} = \{a_1, a_2, ..., a_n\}$ and structural updates as $y_i$ with set $\mathbf{Y} = \{y_1, y_2, ..., y_n\}$,  where $\mathbf{Y}$ comprises all large-scale modifications that can affect an HD map. Finally, $x_i $ is an example set of specific atomic changes that lead to the structural update $y_i$, and $\mathbf{X} = \mathit{P}(\textbf{A})$ the set of all such permutationally invariant subsets of A. The bijective design principle satisfies two conditions:
\begin{itemize}[leftmargin=*,nosep]
  \item \textbf{Surjectivity:} Every structural map update should be explainable in terms of a combination of atomic changes. This ensures coverage.
  \begin{equation}\label{eq:1}\forall  y_i \in \textbf{Y} \ \exists \ x_j \in \textbf{X}: f(x_j) = y_i.\end{equation}
  \item \textbf{Injectivity:} The same structural update should not be represented in two fundamentally different ways. This ensures uniqueness.
   \begin{equation}\label{eq:2}\forall  x_i, x_j \in \textbf{X},\forall  y_i, y_j \in \textbf{Y}: y_j = y_i \Rightarrow x_j = x_i.\end{equation}
\end{itemize}
\begin{table*}[t]
  \centering
  \begin{tabular}{@{}l|ccc|>{\centering\arraybackslash}p{0.9cm} >{\centering\arraybackslash}p{0.9cm} |>{\centering\arraybackslash}p{1.1cm} >{\centering\arraybackslash}p{1.1cm} >{\centering\arraybackslash}p{1.1cm}@{}}
    \toprule
        \multirow{2}{*}{Dataset} & \multirow{2}{*}{Split} & \multirow{2}{*}{Scenes} & \multirow{2}{*}{Avg. duration} & \multicolumn{2}{c|}{HD map} & \multicolumn{3}{c}{Change annotations} \\
        & & & & prior & gt & global & frame & element \\
        \midrule
    nuScenes \cite{nuscenes}& train/val/test & 700/150/150 & 20s &  \ding{55}  & \text{\checkmark} & n.a.  & n.a. & n.a. \\
    Argoverse 2 Sensor \cite{Argoverse2} & train/val/test &
    750/150/100 & 15s & \ding{55}  & \text{\checkmark} & n.a. & n.a. & n.a. \\
    Argoverse 2 TbV \cite{tbv}& train  & 
    799 & 54s & \ding{55} & \text{\checkmark} &  n.a. & n.a. & n.a. \\
     & val  & 
     111 & & \text{\checkmark}  & 
     \ding{55} & \text{\checkmark} & \( \sim \) & \ding{55} \\
     & test  &
     133 & & \text{\checkmark}  & \ding{55} & \ding{55} & \ding{55} & \ding{55} \\
     \midrule
    ArgoTweak (ours) & train/val/test   & 
    697/102/111 & 56s & \text{\checkmark} & \text{\checkmark} &  \text{\checkmark} & \text{\checkmark} & \text{\checkmark} \\
    \bottomrule
  \end{tabular}
  \caption{Comparison of public datasets used in HD mapping, highlighting the availability of priors for training and testing. The last columns detail the presence of change annotations at scenario, \ie, global level, frame-level or element-level. ArgoTweak is the first dataset to complement ground-truth HD maps and sensor data with realistic and real-world map priors, and element-wise annotations.}
  \label{tab:dataset_specs}
\end{table*}
\subsubsection*{Atomic changes and macro-modifications} \ 
To operationalize this framework, we first define the atomic changes $\mathbf{A} = \{a_1, \dots, a_n\}$ as the set of low-level element-wise edit operations on map elements in the Argoverse 2 format \cite{Argoverse2}:
\begin{itemize}[nosep,leftmargin=*]
  \item \textbf{geometry}: modification of element boundaries,
  \item \textbf{markings}: changes to lane markings (type, color),
  \item \textbf{type}: changes to lane semantics (e.g., bus-only),
  \item \textbf{connectivity}: changes to predecessors/successors,
  \item \textbf{insertion} / \textbf{deletion}: creation / removal of map elements.
\end{itemize}

Next, we constrain structural changes $\mathbf{Y}$ to a closed set of interpretable \emph{macro-modifications} $\hat{\mathbf{Y}}$. This design choice is essential: in principle, real-world road layouts can change in unbounded and combinatorially complex ways. Without a limited vocabulary of update types, it would be impossible to maintain a consistent mapping between atomic edits and structural updates. To address this problem, we ask: \emph{how does the local road change, functionally and structurally?} We find that most changes can be meaningfully abstracted into five categories:
\begin{itemize}
    \item \textbf{shape}, \eg, the local road has been widened,
    \item \textbf{appearance}, \eg, the local road has updated markings,
    \item \textbf{function}, \eg, a local road is now reserved for buses,
    \item \textbf{lane graph}, \eg, a new merge has been added,
    \item \textbf{lane number}, \eg, the total lane count in the local road increased/decreased.
\end{itemize}
By defining this interpretable macro space $\hat{\mathbf{Y}}$, we enable a \textbf{soft bijection}: a practical, near-injective mapping from macro-modifications to unique compositions of atomic changes $\mathbf{A}$. While this is an approximation of the full bijectivity between $\mathbf{A}$ and $\mathbf{Y}$, our experiments later demonstrate its consistency and practical usefulness.  

\subsubsection*{Constructing the soft bijection} \
 When constructing the mapping between our atomic change vocabulary $\mathbf{A}$ and constrained macro space $\hat{\mathbf{Y}}$, a central challenge is that many macro-modifications can be represented in multiple ways. For example, a shape change could be expressed via geometry edits to existing elements, or modeled as deleting old segments and inserting new ones. This ambiguity undermines injectivity: the same structural update could correspond to multiple disjoint atomic edit sets. Moreover, connectivity changes create ambiguities -- if map elements are treated as lane graph vertices, adding a new predecessor, for instance, does not necessarily change the role of the edited lane segment itself.

To resolve this, we introduce a disambiguation rule based on the \textbf{underlying road graph}. We represent map elements as \emph{edges} in a non-directional lane graph, where junctions form vertices. Insertions and deletions are used \emph{only} when a change also alters this topology -- e.g., adding a new connection or splitting a lane. If the topology remains unchanged, we express the update as an in-place edit using atomic geometry, marking, or type changes.

This principle guides the design of our mapping matrix $\mathcal{C}$, which encodes which atomic change types contribute to each macro-modification:
\begin{equation}
\mathcal{C} =
{
\begin{array}{c|cccll}
\text{} & \text{geo} & \text{mark} & \text{type} & \text{ins} & \text{del} \\
\hline
\text{shape}       & \checkmark & \text{\ding{55}}  & \text{\ding{55}}  & \checkmark^\ast & \checkmark^\ast \\
\text{appearance}  & \text{\ding{55}}  & \checkmark & \text{\ding{55}}  & \checkmark^\ast & \checkmark^\ast \\
\text{function}    & \text{\ding{55}}  & \text{\ding{55}} & \checkmark & \checkmark^\ast & \checkmark^\ast \\
\text{lane graph}   & \text{\ding{55}}  & \text{\ding{55}}  & \text{\ding{55}}  & \checkmark & \checkmark \\
\text{lane number} & 0 & 0 & 0 & +1 & -1 \\
\end{array}
}
\end{equation}

Here, $\mathcal{C}_{ij}=\checkmark$ indicates that the macro-modification $\hat{y}_i$ is produced by the atomic change $a_j$. For lane number, $\mathcal{C}_{ij}=\pm1$ signals an increase or decrease in the total lane number. The starred entries indicate that insert/delete operations are only used when the lane graph changes too. Details regarding the practical application of the framework and illustrative figures can be found in \ref{sec:bijectivedetails} of the supplementary material. 

\subsection{Building the dataset}
Equipped with our bijective change mapping, we now select the dataset to which we apply our framework. As summarized in \cref{tab:dataset_specs}, existing public datasets for \textit{HD mapping} pair sensor data with ground-truth HD maps but lack outdated priors preceding real-world changes. Such priors are present in \cite{tbv}, a dataset for \textit{HD map change detection}. While \cite{tbv} includes some real-world outdated priors and current sensor data, it does not provide ground truth maps for changed regions, making it unsuitable for developing and evaluating map updating methods.

Another limitation is that real-world stale maps are rare. As a result, real-world priors are only available for validation and testing in \cite{tbv} (\cf \cref{tab:dataset_specs}), whereas for training, the authors propose a rule-based synthetic map modification approach. However, they report a substantial sim2real  gap when models trained on these synthetic priors are evaluated on real-world changes. This challenge is expected to persist when generating updated maps \cite{bateman}.

\definecolor{mplblue}{rgb}{1, 0.6470588235, 0}
\definecolor{mplred}{rgb}{1,0,0}
\definecolor{mpl_green_}{rgb}{0,1,0}
\begin{figure*}[t!]
    \centering
    \begin{tikzpicture}[
        node distance=0.5cm and 0.2cm, 
        every node/.style={font=\footnotesize,rounded corners, line width=0.005cm,               }
        ]
        \node[minimum width=2cm, minimum height=1cm] (sensor) 
{\includegraphics[width=2cm]{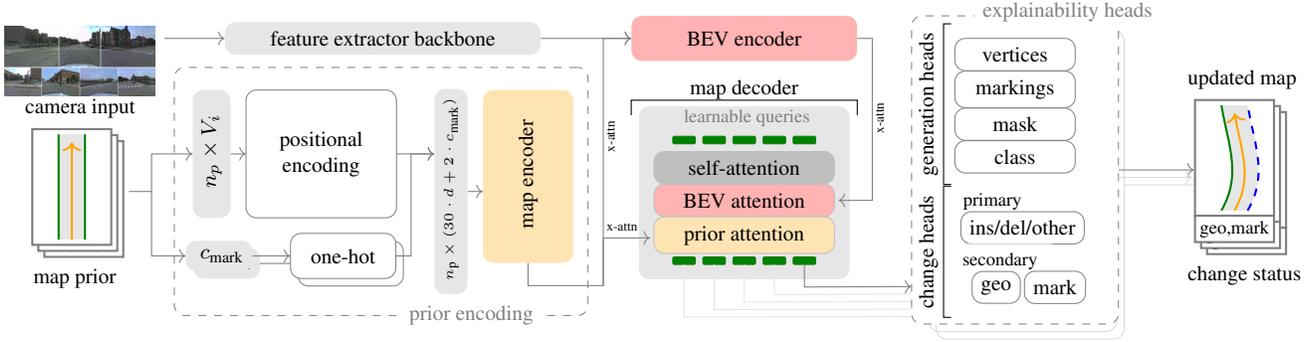}};
        \node[minimum width=2cm, below=-0.2cm of sensor.south]
{camera input};

        \node[fill=gray!20, minimum width= 4.2cm,right=0.8cm of sensor, yshift=0.3cm] (resnet) {\rotatebox{0}{feature extractor backbone}};
        \node[fill=mplred!30,minimum width= 3cm,minimum height=0.6cm,right=1.2cm of resnet] (bev) {\rotatebox{0}{BEV encoder}};

        \node[fill=gray!20, minimum width=2.8cm, minimum height=2.3cm, below=0cm of bev, anchor=north, yshift=-0.6cm] (mapdecoderbox) {};
        \node[ minimum width=2.5cm, minimum height=1.3cm, below=0.5cm of bev, anchor=north, yshift=-0.7cm] (mapdecoder) {map decoder};

        \node[draw=lightgray,fill=lightgray, minimum width = 2.4cm, below=0cm of mapdecoder.north] (self attention) {self-attention};
        \node[draw=lightgray,fill=mplred!30, minimum width = 2.4cm, below=0cm of self attention.south] (bevattention) {BEV attention};
        \node[draw=lightgray, fill=mplblue!30, minimum width = 2.4cm, below=0cm of bevattention.south] (mapattention) {prior attention};

    \tikzstyle{box} = [rounded corners, minimum width=2.5cm, minimum height=0.8cm, draw=gray, text=gray, font=\large]
    \tikzstyle{colorbox} = [rounded corners, minimum width=1.6cm, fill=white, draw=gray, text=black]
    \tikzstyle{chips} = [ rounded corners=0.02cm ,minimum width=0.3cm, minimum height=0.1cm, inner sep=0pt, 
    outer sep=0pt,draw]
        \definecolor{mpl_green}{RGB}{0,128,0}
    \node[rounded corners=0.04cm, minimum width = 2.3cm, minimum height=0.18cm, below= 0.05cm of mapdecoder.south, anchor=north] (querybox1){{}};
    \node[rounded corners=0.04cm, minimum width = 2.3cm, minimum height=0.18cm, above= 0.05cm of mapdecoder.north, anchor=south] (querybox2){{}};
    \node[chips, draw=mpl_green, fill=mpl_green, below= 0.1cm of mapdecoder.south] (query1){{}};
    \node[chips, draw=mpl_green, fill=mpl_green, left= 0.1cm of query1] (query2){{}};
    \node[chips, draw=mpl_green, fill=mpl_green, left= 0.1cm of query2] (query3){{}};
    \node[chips, draw=mpl_green, fill=mpl_green, right= 0.1cm of query1] (query4){{}};
    \node[chips, draw=mpl_green, fill=mpl_green, right= 0.1cm of query4] (query5){{}};
    
    \node[chips, draw=mpl_green, fill=mpl_green, above= 0.1cm of mapdecoder.north] (query6){{}};
    \node[chips, draw=mpl_green, fill=mpl_green, left= 0.1cm of query6] (query7){{}};
    \node[chips, draw=mpl_green, fill=mpl_green, left= 0.1cm of query7] (query8){{}};
    \node[chips, draw=mpl_green, fill=mpl_green, right= 0.1cm of query6] (query9){{}};
    \node[chips, draw=mpl_green, fill=mpl_green, right= 0.1cm of query9] (query10){{}};
    \node[above=0.02cm of query6.north, minimum width=3cm, text=gray] (learnableq){\rotatebox{0}{ \scriptsize learnable queries}};
    \draw[] 
        ([yshift=-0.2cm]learnableq.north west) -- ([]learnableq.north west) -- 
        ([
        ]learnableq.north east) -- ([yshift=-0.2cm]learnableq.north east);
    \node[above=-0.1cm of learnableq.north, minimum height = 0.1cm]  {map decoder};
    
    \foreach \i in {1,2} {
        \node[draw=gray!30, minimum width=2.5cm, minimum height=4.1cm,
              right=3.6cm+\i*0.1cm of bev.north, yshift=-\i*0.1cm, anchor=north] (rectangle\i) {};
    }
    \draw[->, draw=gray, draw=gray!20] (query4.south) -- ([yshift=-0.4cm] query4.south) -- ([yshift=0.5cm] rectangle1.south west);
        \draw[->, draw=gray, draw=gray!20] (query1.south) -- ([yshift=-0.5cm] query1.south) -- ([yshift=0.5cm] rectangle2.south west);
        \draw[->, draw=gray, draw=gray!20] (query2.south) -- ([yshift=-0.6cm] query2.south) -- ([yshift=0.4cm] rectangle2.south west);
        \draw[->, draw=gray, draw=gray!20] (query3.south) -- ([yshift=-0.7cm] query3.south) -- ([yshift=0.3cm] rectangle2.south west);
       
    \node[draw=gray, dashed, minimum width=2.75cm, minimum height=4.1cm,right=3.6cm of bev.north, anchor=north,fill=white] (rectangle) {};
    \node[text=gray,above=of rectangle.north, fill=white, yshift=-0.7cm, xshift=0.7cm] (rectanglelabel) {explainability heads};
   
    \node[colorbox, below= 0.3cm of rectangle.north, xshift=0cm] (centerline){{vertices}};
    \node[colorbox, below= 0cm of centerline.south] (offset){{markings}};;
    \node[colorbox, below= 0cm of offset] (linetype){{mask}};;
    \node[colorbox, below= 0cm of linetype] (topology){{class}};
    \node[colorbox,draw=white,below= 0.15cm of topology, minimum width=0.1cm, xshift=-0.3cm] (pm){\scriptsize{primary}};
    \node[colorbox, below= -0.1cm of pm, minimum width=0.1cm, xshift=0.4cm] (insdelother){{ins/del/other}};
    \node[colorbox,draw=white!0,below= 0cm of insdelother,xshift=-0.3cm] (pm2){\scriptsize{secondary}};
    \node[colorbox, minimum width=0.1cm, fill=white,below= -0.1cm of pm2,xshift=-0.05cm] (geo){geo};
    \node[colorbox, minimum width=0.1cm,fill=white,right= 0.05cm of geo] (mark){mark};

    \node[right= -0.6cm of offset.south west, minimum width=0.5cm,yshift=0.05cm, minimum height=1.8cm] (predictors){\rotatebox{90}{generation heads}};
    \node[right= -0.4cm of pm2.north west, minimum width=0.5cm, yshift=-0.1cm] (changeheads){\rotatebox{90}{change heads}};

    \draw[] ([xshift=0.3cm] predictors.south) -- ([xshift=0.2cm] predictors.south) --
    ([xshift=0.2cm] predictors.north)--
    ([xshift=0.3cm] predictors.north);
    \draw[] ([xshift=0.3cm] changeheads.south) -- ([xshift=0.2cm] changeheads.south) --
    ([xshift=0.2cm] changeheads.north)--
    ([xshift=0.3cm] changeheads.north);

        \node[draw=gray, dashed,rounded corners, minimum width=5.5cm, minimum height=3.3cm, below= of sensor, xshift=4cm, yshift=1cm] (tokenizer) {};
        \node[fill=white,rounded corners, text=gray, below= of tokenizer.south, xshift=1.2cm, yshift=0.75cm] (tokenizerlabel) {prior encoding };
    \definecolor{ribbonColor}{RGB}{230, 230, 230}
    \definecolor{mpl_green}{RGB}{0,128,0}
    \definecolor{mpl_green_}{RGB}{255, 165, 0}
    \definecolor{mpl_blue}{RGB}{0, 0, 255}

    \foreach \i in {0,1,2} {
        \begin{scope}[shift={(tokenizer.west)}, xshift=-1.2cm-\i*0.08cm, yshift=-0.1cm+\i*0.08cm, scale=0.35]
        
            \draw[ gray, thick, fill=white]  (-1.5,-2.2) rectangle (1.5,2.2);

            \fill[ribbonColor] 
                (-0.5,-2) to[out=90,in=90] (-0.5,2)
                -- (0.5,2) to[out=90,in=90] (0.5,-2) -- cycle;

            \draw[thick, mpl_green] 
                (-0.5,-2) to[out=90,in=-90] (-0.5,2);
            \draw[thick, mpl_green] 
                (0.5,-2) to[out=90,in=-90] (0.5,2);

            \draw[->, draw=gray, thick, mpl_green_] 
                (0,-2) to[out=90,in=-90] (0,1.7);

        \end{scope}
    }
    \node[left=0.65cm of tokenizer.west, yshift=-1.15cm] 
{map prior};
        \node[fill=gray!20, minimum height=1.7cm] (input) at ([xshift=0.5cm, yshift=0.5cm] tokenizer.west) {\rotatebox{90}{ $n_p\times V_i$}};
        
        \node[fill=gray!20, below=0.4cm of input.south, xshift=0.2cm] (input_3)  {\rotatebox{00}{ $c_\text{mark}$}};
        \node[fill=gray!20, below=0.3cm of input.south,, xshift=0.1cm] (input_2)  {\rotatebox{00}{ $c_\text{mark}$}};

        \node[draw=gray, minimum width=2cm, minimum height=1.7cm, right=0.2cm of input] (mlp1) 
{
    \begin{tabular}{c}
        positional \\
        encoding
    \end{tabular}
};

        \node[draw=gray, minimum width=1.4cm,fill=white, minimum height=0.6cm, right=0.6cm of input_2, yshift=-0.1cm] (mlp_shared_3) {one-hot};
        \draw[->, draw=gray] (input_3.east) |- (mlp_shared_3.west);
        \node[draw=gray, minimum width=1.4cm,fill=white, minimum height=0.6cm, right=0.5cm of input_2] (mlp_shared) {one-hot};

        \node[fill=gray!20, minimum height=2.2cm, right=0.5cm of mlp1, yshift=-0.5cm] (point_tokens) {\rotatebox{90}{ \tiny$
n_{\text{p}} \times (30 \cdot d + 2 \cdot c_\text{mark})
$}};

        \node[ fill=mplblue!30, minimum width=1.2cm, minimum height=2.3cm, right=0.2cm of point_tokens, yshift=0.2cm] (hier_tokens) {\rotatebox{90}{ map encoder}};
        \draw[->, draw=gray] (rectangle.east) -- ([xshift=1cm]rectangle.east);
        \draw[->, draw=gray, draw=gray!30] (rectangle1.east) -- ([xshift=1cm]rectangle1.east);
        \draw[->, draw=gray, draw=gray!30] (rectangle2.east) -- ([xshift=1cm]rectangle2.east);

        \foreach \i in {0,1,2} {
        \begin{scope}[shift={(rectangle.east)}, xshift=1.7cm-\i*0.08cm, yshift=0.cm+\i*0.08cm, scale=0.35]

   \draw[gray, thick, fill=white] (-1.5,-2.2) rectangle (1.5,2.2);
   \draw[gray, thick, fill=white] (-1.5,-3.2) rectangle (1.5,-2.2);
\node at (0,-2.7) {\scriptsize geo,mark};

    \fill[ribbonColor] 
        (-0.5,-2) to[out=60,in=-80] (-0.5,2)
        -- (0.5,2) to[out=-80,in=60] (0.5,-2) -- cycle;

    \draw[thick, mpl_green] 
        (-0.5,-2) to[out=60,in=-80] (-0.5,2);
    \draw[thick, dashed, mpl_blue] 
        (0.5,-2) to[out=60,in=-80] (0.5,2);
    \draw[->, draw=gray,thick, mpl_green_] 
        (0,-2) to[out=60,in=-80] (0,1.7);
        \end{scope}}

        \node[right=of rectangle.east, yshift=-1.4cm,xshift=0.6cm] 
{change status};
        \node[right=of rectangle.east, yshift=1.2cm,xshift=0.6cm] 
{updated map};
        \draw[->, draw=gray] (input.east) -- (mlp1.west);
        
        \draw[->, draw=gray] (input_2.east) |- (mlp_shared.west);
        
        \draw[->, draw=gray] (mlp1.east) |- ([yshift=0.5cm] point_tokens.west) ;
        \draw[->, draw=gray] (mlp_shared.east) --([xshift=0.2cm] mlp_shared.east)|- ([yshift=0.5cm] point_tokens.west) ;
        \draw[->, draw=gray] (point_tokens.east) -- ([yshift=-0.2cm] hier_tokens.west);
         \draw[->, draw=gray] ([yshift=0.3cm] sensor.east) -- (resnet);
         \draw[->, draw=gray] (resnet) -- (bev);
        
        \draw[->, draw=gray] (hier_tokens.south) -- ([yshift=-0.3cm] hier_tokens.south)-- ([yshift=-0.3cm, xshift=1cm] hier_tokens.south) |- node[midway, yshift=0.15cm, xshift=0.25cm] {\rotatebox{0}{\tiny x-attn}} ( [yshift=-0.5cm] mapdecoder.west);
        \draw[->, draw=gray] ([yshift=-0.3cm, xshift=1cm] hier_tokens.south)  
        --  ([yshift=2.6cm, xshift=1cm] hier_tokens.south)    |- node[midway, yshift=-1.3cm, xshift=0.1cm] {\rotatebox{90}{\tiny x-attn}} ( [] bev.west);
      
        \draw[->, draw=gray] (bev.east) -- ([xshift=0.2cm] bev.east)  |- node[midway, yshift=1.1cm, xshift=0.1cm] {\rotatebox{90}{\tiny x-attn}}( [yshift=-0cm] mapdecoder.east);
        
        \draw[->, draw=gray] ([xshift=-.6cm, yshift=0] tokenizer.west) --([xshift=-.3cm, yshift=0] tokenizer.west)|- (input.west);
        \draw[->, draw=gray] ([xshift=-.3cm, yshift=0] tokenizer.west)|- (input_2.west);

        \draw[->, draw=gray] (query5.south) -- ([yshift=-0.3cm] query5.south) -- ([yshift=0.5cm] rectangle.south west);

    \end{tikzpicture}
    
    \caption{Network architecture overview. A  BEV encoder and a prior encoder extract features from camera input and map prior, while the map decoder predicts both updated map elements and their change status (e.g., insertion, deletion, geometry/marking edits) through explainable multi-head training. See \cref{sec:architecturedetails} for implementation details.} \label{fig:architecture}
\end{figure*}
To address these limitations, we construct a high-quality training set by introducing realistic structural modifications to ground-truth maps within our bijective change mapping framework. This approach ensures full control over the nature and distribution of changes, eliminating the need to mine rare real-world priors while maintaining high realism. For testing, we leverage the real-world priors and up-to-date sensor data from the validation split of \cite{tbv}, and annotate the complementing ground truth maps. We use the former validation as our test set, as global and frame-wise annotations are not publicly available for the original test set, making change localization challenging.

Finally, to systematically analyze a potential sim2real  gap, we reorganize the original training split. Maps from Washington, D.C., are designated exclusively for validation to prevent geographical data leakage \cite{lilja, StreamMapNet}. This setup allows us to measure the sim2real  gap by comparing performance on realistic but manually created validation maps against the real-world test set.

\subsection{Additional dataset refinements}\label{sec:refinements}
We refine the base dataset \cite{tbv} with several improvements. Following \cite{LaneSegNet}, we apply an OpenLane-V2-inspired \cite{wang2023openlanev2} merging process to prior and updated maps, resolving unnecessary breakpoints in lane segments. The search boundaries for element merging include lane graph changes, changes to the lane properties and the tracked change status in terms of atomic changes. Additionally, we unify pedestrian crossing edges, previously defined in both clockwise and counterclockwise orientations. While our priors and ground truth maps are in 2D, z-coordinates can be sampled from the base dataset’s ground height annotations as needed. Detailed dataset statistics can be found in \ref{sec:datasetstats} of the supplementary material.
\section{Explainable prior-aided mapping}
\label{sec:method}
\subsubsection*{Task definition} \ \textit{At timestamp $t$, given current sensor data and prior map $M_{prior}$, the goal is to estimate whether $M_{prior}$ is in agreement with current sensor data, detect changed elements and update them accordingly to produce the ground truth map $M_{gt}$. This includes detecting changes in lane markings and types, insertions, deletions, and geometric modifications at an element level.}

\subsection{Network architecture} \label{sec:architecturedetails}
To serve as a baseline for future research and a vehicle for our evaluation, we propose a flexible map updating scheme that can operate at different levels of explainability: without explicit change assessment (\ie, not explainable), with a binary change detection head (\ie changed/unchanged), or with a full explainability module that attributes specific atomic changes to individual map elements.
We adopt LaneSegNet \cite{LaneSegNet} as our backbone due to its lane-segment-based formulation, which aligns well with our bijective mapping framework. We extend the backbone by adding a map prior encoder and explainability heads (\cref{fig:architecture}). 
\subsubsection*{Prior encoding} \ \label{sec:priorencoding}
We use the map prior encoding scheme proposed in \cite{smerf}, which we previously demonstrated to be compatible with LaneSegNet~\cite{LaneSegNet} in our earlier work~\cite{ExelMap}. Our prior includes 10 2D-points for left and right boundary and centerline, class labels (pedestrian crossing or lane), and semantic attributes (left/right lane line markings). We extract the geometric representation $V$ for each of the \( n_{\text{prior}} \) elements in the prior map, \[
V = \{V_\text{left}, V_\text{right}, V_\text{center}\} = \{(x_i, y_i)\}_{i=1}^{30}.
\]
Lane line markings are encoded using a one-hot scheme for all $7$ marking types $c_\text{mark}$. Lane marking color information is not used in the present configuration. The same holds for lane type, as none of the current map generation networks  detect bus or bike lane separately. The encoded coordinates and boundary types are concatenated into a polyline sequence of shape $
n_{\text{prior}} \times (30 \cdot d + 2 \cdot c_\text{mark}),
$
where \( d \) is the positional embedding dimension.

Once encoded, the prior can be incorporated into the map generation pipeline in two ways. The first approach, used in, \eg, \cite{smerf, ExelMap}, treats the prior as an additional modality in cross-attention alongside BEV-features. However, bandwidth limitations potentially hinder full integration \cite{bateman}. The second approach replaces fixed hierarchical queries with prior tokens, refining them into a posterior map through decoder layers \cite{bateman, MtM, m3tr}. While strategy 2 can improve stability, we found strategy 1 to be more flexible and effective for integrating larger changes. 

\subsubsection*{Explainability heads} \ \label{sec:eh}
In our previous work \cite{ExelMap}, we proposed the idea of using multi-task learning with change assessment heads. Since that approach was limited to insertions and deletions, we redesign the heads to handle a more complex setting with diverse atomic changes.

Our key insight is that some change categories are mutually exclusive, while others can co-occur. Inserted or deleted elements cannot meaningfully undergo further changes, whereas geometry and lane marking modifications can happen simultaneously on a single element. Thus, we introduce a two-stage assessment: a primary multi-class classification head determines the element's status from mutually exclusive labels 
\{\text{No Change}, \text{Insertion}, \text{Deletion}, \text{Other}\}.
If classified as "Other," secondary binary heads can be considered for geometric and lane marking modifications. This modular design allows swift integration of additional change categories in the future.

With these three heads, our final loss is 
\begin{equation}
\begin{aligned}
\mathcal{L}& =  \ \lambda_{\text{vec}} \mathcal{L}_{\text{vec}} + \lambda_{\text{seg}} \mathcal{L}_{\text{seg}} + \lambda_{\text{cls}} \mathcal{L}_{\text{cls}} + \lambda_{\text{type}} \mathcal{L}_{\text{type}} \\
& + \lambda_{\text{cd,prim.}} \mathcal{L}_{\text{cd,prim.}} + \sum_{i\in[\text{geo},\text{mark}]}\lambda_{\text{cd,sec.}}^i \mathcal{L}_{\text{cd,sec.}}^i,
\end{aligned}
\end{equation}
where \( \mathcal{L}_{\text{seg}} = \lambda_{\text{ce}} \mathcal{L}_{\text{ce}} + \lambda_{\text{dice}} \mathcal{L}_{\text{dice}} \) is the segmentation loss from \cite{LaneSegNet}, and the loss weights are defined as \(\lambda_{\text{vec}} = 0.025\), \(\lambda_{\text{seg}} = 3.0\), \(\lambda_{\text{ce}} = 1.0\), \(\lambda_{\text{dice}} = 1.0\), 
\(\lambda_{\text{cls}} = 1.5\), \(\lambda_{\text{type}} = 0.01\), \(\lambda_{\text{cd,primary}} = \lambda_{\text{cd,secondary}}^i =  0.5\). We use cross-entropy loss for the primary and Focal Loss \cite{focalloss} for the secondary heads.

\section{Metric}\label{sec:metric}
Complementing our dataset and explainable model, we introduce a fine-grained evaluation protocol as the third pillar to systematically assess both stability in unchanged regions and responsiveness to updates.
Inspired by \cite{m3tr} and \cite{ExelMap}, we propose a change aware dual-metric framework that comprises a coarse detection accuracy (mACC) and fine-grained map generation average precision (mAPC).
\subsubsection*{Fine-grained mAPC} \
Given the set of change categories \(\mathcal{C}\) of size \(|\mathcal{C}| = C\), we evaluate each predicted map element \(\hat{V}\) against its ground-truth counterpart $V$ \textit{only if} their predicted and true change status match, \(\hat{c}_V = c_V\). The change-aware lane segment distance is defined by adapting \cite{LaneSegNet} to 
\begin{align}
D_{ls}^{c}(V,& \hat{V}) = \frac{1}{2} \Big[ \text{Chamfer}([V_\text{left}, V_\text{right}], 
[\hat{V}_{\text{left}}, \hat{V}_{\text{right}}]) \notag \\ 
&\quad + \text{Fr\'echet}(V_{\text{center}}, \hat{V}_{\text{center}}) \Big] 
\otimes \mathbf{1}\{\hat{c}_{V} = c_{V}\}.
\end{align}
We compute the average precision per class ($\text{AP}_c$) at distance thresholds \(\{1.0, 2.0, 3.0\}\)m. For non-directional pedestrian crossings, we use Chamfer distance at \(\{0.5, 1.0, 1.5\}\)m. The class-wise $\text{mAP}_c$ combines lane segments and pedestrian crossings,
\begin{equation}
    \text{mAP}_c= \frac{1}{2} (\text{AP}_c^\text{ls} + \text{AP}_c^\text{pc}).
\end{equation}
Finally, the overall class-aware mean average precision is
\begin{equation}
    \text{mAPC} = \frac{1}{C} \sum_{c \in \mathcal{C}} \text{mAP}_c.
\end{equation}
This ensures equal weighting across object and change types, preventing dominance by over-represented classes.
\subsubsection*{Coarse mACC} \
Extending \cite{tbv} and our prior work~\cite{ExelMap}, we introduce a coarse change detection metric. For each frame, a binary ground-truth label \(y_c \in \{0,1\}\) indicates whether at least one map element changed for change type \(c \in \mathcal{C}\). The model’s prediction \(\hat{y_c} \in \{0,1\}\) follows the same criterion. Class-wise precision and recall are captured via
\begin{equation}
    \text{Acc}_c^{+(-)} = 
    \frac{
     \sum\limits_{i=1}^{N}
    \mathds{1} \{ \hat{y}_{c} = y_{c} \} \cdot 
    \mathds{1} \{ y_{c} = 1 \ (0) \}
    }{
    \sum\limits_{i=1}^{N}
    \mathds{1} \{ y_{c} = 1 \ (0) \}
    }.
\end{equation}
The final accuracy metric is:
\begin{equation}
    \text{mACC} = \frac{1}{C} \sum_{c\in \mathcal{C}} \text{mAcc}_c, 
    \end{equation}with \quad \begin{equation}\text{mAcc}_c = \frac{1}{2}(\text{Acc}_c^{+}+\text{Acc}_c^{-}).
\end{equation}

Evaluating both mACC and mAPC offers several advantages. The coarse mACC offers an initial assessment, with low scores signaling a risk of missing updates, while fine-grained mAPC evaluates element accuracy by change type. High mACC but low mAPC suggests detecting changes without precise localization, whereas poor mACC may indicate an overly conservative model. This multi-tiered evaluation framework facilitates model refinement and establishes a foundation for nuanced comparison.

\section{Experiments}
\label{sec:experiments}
We train our network for 10 epochs with
a batch size of 8 and AdamW optimizer on 8 NVIDIA A10G Tensor Core GPUs. For feature extraction, we employ a pretrained ResNet-50 \cite{resnet}. We use camera as the only sensor input and crop our map size to $50\times50\text{m}^2$ \cite{ExelMap}. While not speed-optimized, our model runs at $\sim$4 FPS on a single NVIDIA A10G GPU. 

\subsection{Map updating without change modelling}\label{sec:syntheticpriorcomparison} 
In this experiment, we examine how different priors influence model performance to highlight the inadequacy of mAP for evaluating prior-aided mapping. We train our network on ArgoTweak, and on synthetically generated priors inspired by established prior-generation methods \cite{bateman, MtM, m3tr, PriorDrive, tbv, ExelMap}. We consider three types of priors:
\begin{itemize}
    \item \textbf{Continuous modifications}: We add Gaussian noise ($\mu=0, \sigma=0.5$) to all vertices of the ground truth map.
    \item \textbf{Discrete modifications}: We randomly delete or shift entire map elements, with a Gaussian drift ($\mu=0, \sigma=0.5$) and probabilities $p_\text{del} = p_\text{shift} = 0.2$.
    \item \textbf{Rule-based modifications}: Using the approach in \cite{tbv}, rule-based, scripted edits such as pedestrian crossing insertions and deletions, bike lane additions, and lane marking changes are generated (\cf \ref{sec:simpletbv}). 
\end{itemize}
To ensure fair comparison, we use neither change annotations nor change assessment heads, because synthetic priors -- especially those generated through noise perturbations -- cannot meaningfully be described by atomic changes.
\begin{figure}
    \centering
     \label{fig:syntheticpriorcomparison}

    \begin{subfigure}{0.13\textwidth}
    \centering
    \fboxsep=0pt
    \rotatebox{-90}{\adjustbox{padding=0.5pt}{\includegraphics[width=\textwidth]{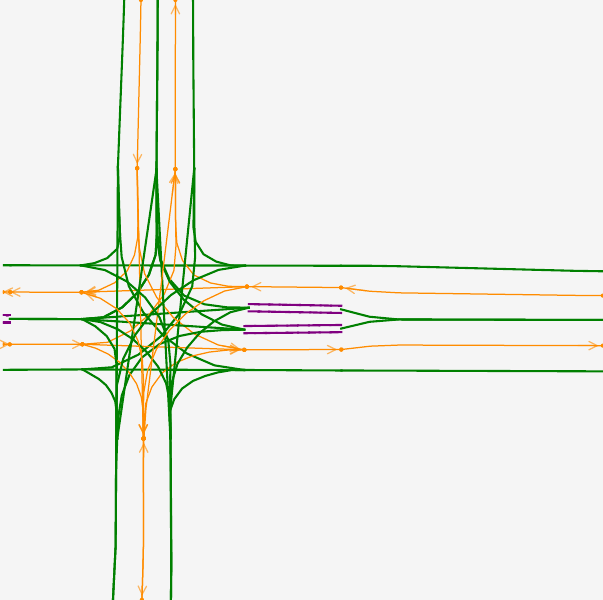}}}
    \caption{map prior} 
\end{subfigure}
\hfill
\hspace{0.5\fboxsep}
   \textcolor{gray}{\rule{0.2mm}{30mm} }
\hfill
    \begin{subfigure}{0.13\textwidth}
        \centering
        \fboxsep=0pt
        \rotatebox{-90}{\adjustbox{padding=0.5pt}{\includegraphics[width=\textwidth]{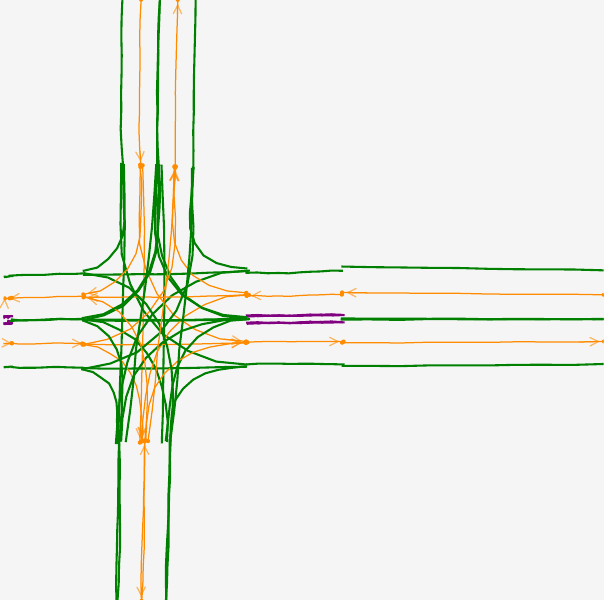}}}
        \caption{discrete}
        \label{fig:discrete}
    \end{subfigure}
    \begin{subfigure}{0.13\textwidth}
        \centering
        \fboxsep=0pt
         \rotatebox{-90}{\adjustbox{padding=0.5pt}{\includegraphics[width=\textwidth]{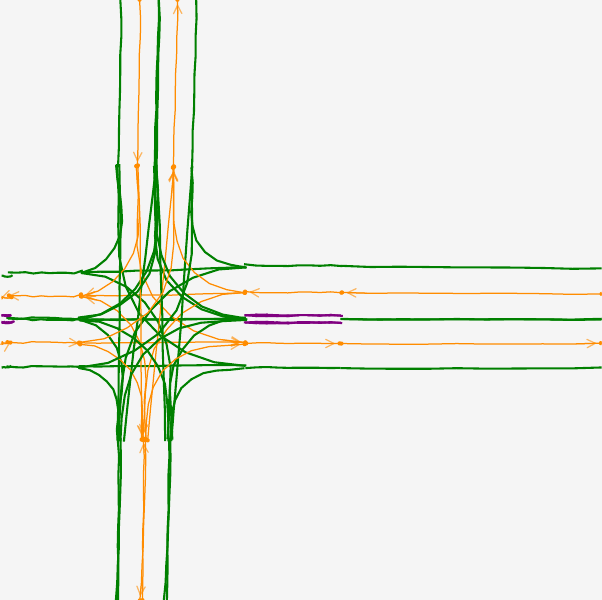}}}
        \caption{continuous}
        \label{fig:continuous}
    \end{subfigure}

\vspace{-\fboxsep}
    \begin{subfigure}{0.13\textwidth}
        \centering
        \fboxsep=0pt
        \rotatebox{-90}{\adjustbox{padding=0.5pt}{\includegraphics[width=\textwidth]{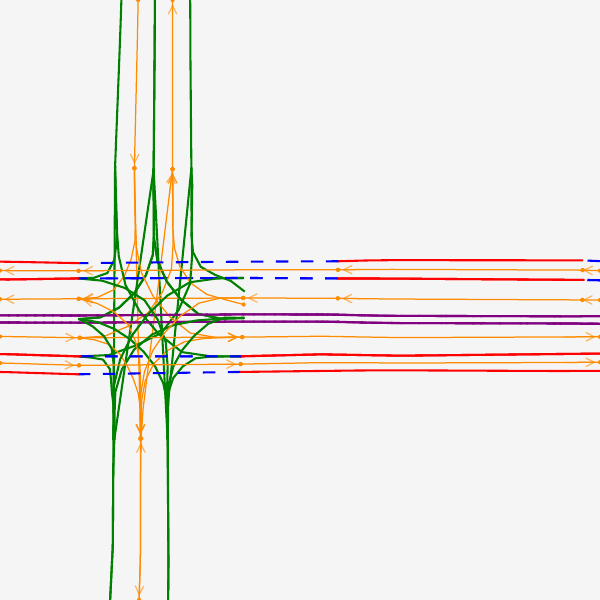}}}
        \caption{ground truth}
        \label{fig:gt}
    \end{subfigure}
   \hfill
   \hspace{0.5\fboxsep}
   \textcolor{gray}{\rule{0.2mm}{31mm} }
   \hfill
    \begin{subfigure}{0.13\textwidth}
        \centering
        \fboxsep=0pt
          \rotatebox{-90}{\adjustbox{padding=0.5pt}{\includegraphics[width=\textwidth]{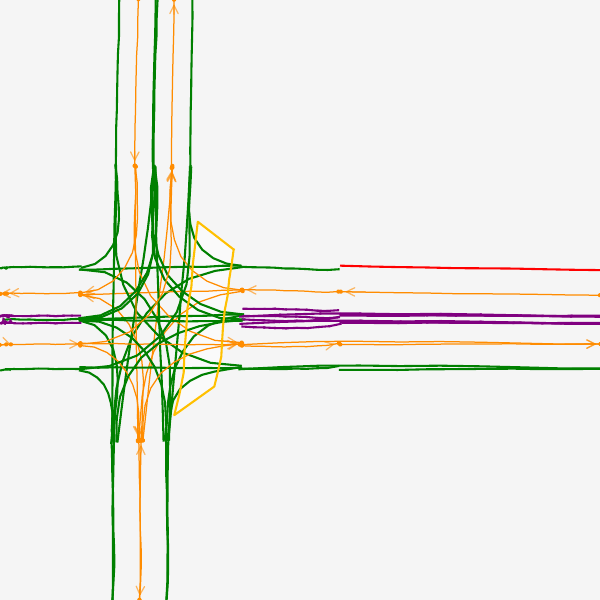}}}
        \caption{rule-based}
        \label{fig:rulebased}
    \end{subfigure}
    \begin{subfigure}{0.13\textwidth}
        \centering
        \fboxsep=0pt
        \rotatebox{-90}{\adjustbox{padding=0.5pt}{\includegraphics[width=\textwidth]{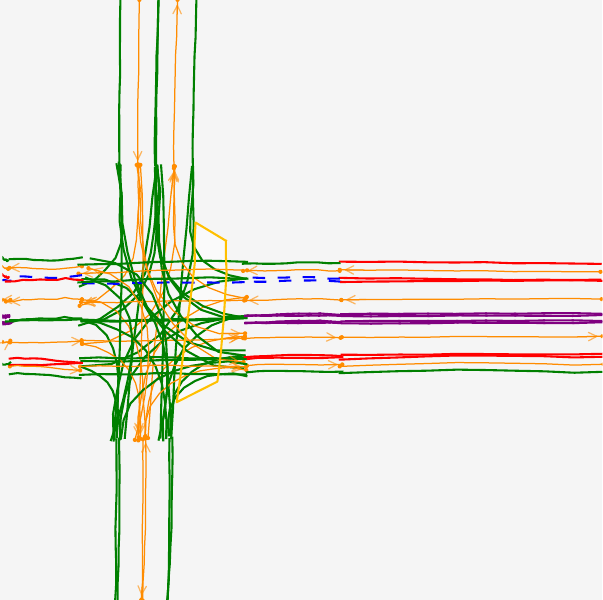}}}
        \caption{ArgoTweak}
        \label{fig:argotweak}
    \end{subfigure}

    \caption{
Qualitative comparison of model outputs with different priors. Only the ArgoTweak-trained network captures complex road updates, while models trained on synthetic priors produce limited or overfit edits. However, these differences are not reflected in mAP scores (\cref{tab:results}), illustrating a limitation of standard metrics.}

    \label{fig:siximages}
\end{figure}
\begin{table}[t]
\centering
\begin{tabular}{@{}llll@{}}
\toprule
Map prior & $\text{AP}_\text{ls}$ & $\text{AP}_\text{pc}$ & mAP \\ 
\midrule
no prior (baseline)& 32.9 & 45.9 & 39.4 \\
continuous modifications& 71.6 & 75.5 & 73.5 \\
discrete modifications& 71.0 & 71.9 & 71.5 \\
rule-based editing \cite{tbv} & \underline{74.2} & \underline{79.3} &  \underline{76.7} \\
\midrule
ArgoTweak & \textbf{75.8} & \textbf{79.6} & \textbf{77.7} \\ 
\bottomrule
\end{tabular}
\caption{
Quantitative comparison of models trained on different priors. Despite strong qualitative variation across outputs   (\cref{fig:siximages}), this is not reflected in the similar mAP scores, highlighting the need for change-aware evaluation.}

\label{tab:results}
\end{table}
While the mAP for our ArgoTweak-trained model is only slightly higher than when trained on synthetic priors (a ~1\% gain, \cref{tab:results}), this result is by design and central to our argument. The purpose of this experiment is not to showcase large mAP improvements, but to demonstrate that mAP fails to reflect meaningful differences in model behavior.
Despite similar mAP values across all priors, manual inspection reveals stark qualitative differences (\cref{fig:siximages}). Specifically, the ArgoTweak-trained model captures complex map updates; models trained on rule-based priors tend to overfit to lane marking changes; and noise-based priors only support minor geometric corrections.  Notably, models trained on synthetic priors show greater stability in unchanged regions.

These observations underscore a critical limitation: Without a notion of change, existing metrics like mAP collapse very different behaviors into nearly identical scores. This masks key tradeoffs in stability vs. adaptability, and offers no actionable guidance for improving model responsiveness to updates. The present findings reinforce that beyond manual inspection and with only mAP as a metric we are effectively blind to meaningful model differences. 

\pgfdeclarepatternformonly{staggereddots}{\pgfqpoint{-1pt}{-1pt}}{\pgfqpoint{3pt}{3pt}}{\pgfqpoint{2pt}{2pt}}{
    \pgfpathcircle{\pgfqpoint{0.5pt}{0.5pt}}{0.3pt} 
    \pgfusepath{fill}
    \pgfpathcircle{\pgfqpoint{1.5pt}{1.5pt}}{0.3pt} 
    \pgfusepath{fill}
}

\definecolor{mplblue}{rgb}{1, 0.6470588235, 0}
\definecolor{mplred}{rgb}{1,0,0}
\subsection{Sim2real gap assessment} \label{sec:sim2real}
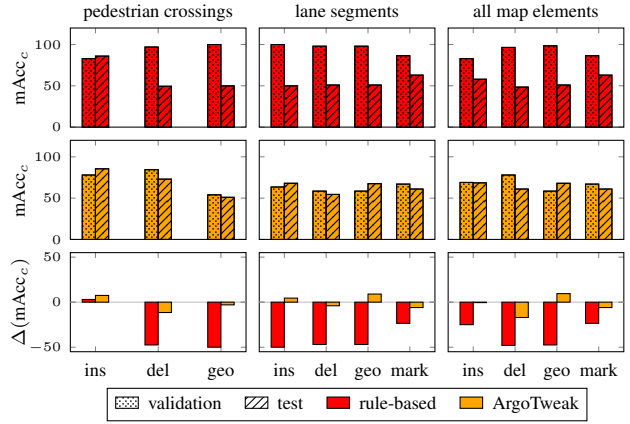
\begin{figure}  \label{fig:sim2real}
    \centering
    \begin{tikzpicture}
        
       \begin{groupplot}[
    group style={group size=3 by 3, horizontal sep=5pt, vertical sep=5pt}, 
    ybar,
    symbolic x coords={ins,del,geo,mark},
    xtick=data,
    ymin=0, ymax=120,
    width=0.47\linewidth, 
    height=2.9cm,
    legend style={font=\tiny, draw=black, fill=white}, 
    enlarge x limits=0.2,
    trim axis left, trim axis right,
    tick style={inner sep=2pt, major tick length=2pt},
    tick align=inside,
    every axis x label/.style={yshift=-2pt, baseline} 
]

\nextgroupplot[
    title={pedestrian crossings},
    title style={font=\scriptsize, yshift=-7pt}, 
    xlabel={}, xticklabels={}, 
    ylabel={$\text{mAcc}_c$}, 
    ylabel style={font=\scriptsize, yshift=-16pt}, 
    font=\tiny, bar width=5pt,
]
 \addplot[fill=mplred,bar shift=-2.5pt, postaction={pattern=staggereddots, pattern color=black}] 
    coordinates {(ins,83.0) (del,97.0) (geo,100.0) };
\addplot[fill=mplred, bar shift=2.5pt, postaction={pattern=north east lines, pattern color=black}] 
    coordinates {(ins,86.0) (del,49.5) (geo,50.0) };
    
\nextgroupplot[title={lane segments}, title style={font=\scriptsize, yshift=-7pt}, 
    xlabel={}, xticklabels={}, ylabel={}, yticklabels={}, font=\scriptsize, bar width=5pt]
\addplot[fill=mplred, bar shift=-2.5pt,postaction={pattern=staggereddots, pattern color=black}] 
    coordinates {(ins,100.0) (del,98.0) (geo,98.0) (mark,86.5)};
\addplot[fill=mplred, bar shift=2.5pt, postaction={pattern=north east lines, pattern color=black}] 
    coordinates {(ins,50.0) (del,51.0) (geo,51.0) (mark,63.0)};

\nextgroupplot[title={all map elements}, title style={font=\scriptsize, yshift=-7pt}, 
    xlabel={}, xticklabels={}, ylabel={}, yticklabels={}, font=\scriptsize, bar width=5pt]
\addplot[fill=mplred, bar shift=-2.5pt,postaction={pattern=staggereddots, pattern color=black}] 
    coordinates {(ins,83.0) (del,96.5) (geo,98.5) (mark,86.5)};
\addplot[fill=mplred, bar shift=2.5pt, postaction={pattern=north east lines, pattern color=black}] 
    coordinates {(ins,58.0) (del,48.5) (geo,51.0) (mark,63.0)};

\nextgroupplot[xlabel={}, xticklabels={}, ylabel={$\text{mAcc}_c$}, 
    ylabel style={font=\scriptsize, yshift=-16pt}, font=\tiny, bar width=5pt,]
\addplot[fill=mplblue, bar shift=-2.5pt, postaction={pattern=staggereddots, pattern color=black}] 
    coordinates {(ins,78.0) (del,84.5) (geo,54.0) };
\addplot[fill=mplblue, bar shift=2.5pt, postaction={pattern=north east lines, pattern color=black}] 
    coordinates {(ins,85.5) (del,73.0) (geo,51.0) };

\nextgroupplot[xlabel={}, xticklabels={}, ylabel={}, yticklabels={}, font=\scriptsize, bar width=5pt]
\addplot[fill=mplblue, bar shift=-2.5pt, postaction={pattern=staggereddots, pattern color=black}] 
    coordinates {(ins,63.5) (del,58.5) (geo,58.5) (mark,67.0)};
\addplot[fill=mplblue, bar shift=2.5pt, postaction={pattern=north east lines, pattern color=black}] 
    coordinates {(ins,68.0) (del,54.5) (geo,67.5) (mark,61.0)};

\nextgroupplot[xlabel={}, xticklabels={}, ylabel={}, yticklabels={}, font=\scriptsize, bar width=5pt]
\addplot[fill=mplblue, bar shift=-2.5pt, postaction={pattern=staggereddots, pattern color=black}] 
    coordinates {(ins,69.0) (del,78.0) (geo,58.5) (mark,67.0)};
\addplot[fill=mplblue, bar shift=2.5pt, postaction={pattern=north east lines, pattern color=black}] 
    coordinates {(ins,68.5) (del,61.0) (geo,68.0) (mark,61.0)};

\nextgroupplot[xticklabels={\strut ins, \strut del, \strut geo}, 
    ylabel={$\Delta(\text{mAcc}_c)$}, 
    ymin=-55, ymax=55,
    ylabel style={font=\scriptsize, yshift=-16pt}, 
    yticklabel style={font=\tiny}, xticklabel style={font=\scriptsize}, 
    font=\scriptsize, bar width=5pt,extra y ticks={0},
extra y tick labels={},
extra y tick style={grid=major, thin, black}]
\addplot[fill=mplred, bar shift=-2.5pt] 
    coordinates {(ins,3) (del,-47.5) (geo,-50) };
\addplot[fill=mplblue, bar shift=2.5pt] 
    coordinates {(ins,7.5) (del,-11.5) (geo,-3) };

\nextgroupplot[ylabel={}, yticklabels={}, font=\scriptsize, bar width=5pt,  ymin=-55, ymax=55,xticklabels={\strut ins, \strut del, \strut geo, \strut mark},
extra y ticks={0},
extra y tick labels={},
extra y tick style={grid=major, thin, black}]
\addplot[fill=mplred, bar shift=-2.5pt] 
    coordinates {(ins,-50) (del,-47) (geo,-47) (mark,-23.5)};
\addplot[fill=mplblue, bar shift=2.5pt] 
    coordinates {(ins,4.5) (del,-4) (geo,9) (mark,-6)};

\nextgroupplot[ylabel={}, yticklabels={}, font=\scriptsize, bar width=5pt,  ymin=-55, ymax=55,xticklabels={\strut ins, \strut del, \strut geo, \strut mark},extra y ticks={0},
extra y tick labels={},
extra y tick style={grid=major, thin, black}]
\addplot[fill=mplred, bar shift=-2.5pt] 
    coordinates {(ins,-25) (del,-48) (geo,-47.5) (mark,-23.5)};
\addplot[fill=mplblue, bar shift=2.5pt] 
    coordinates {(ins,-0.5) (del,-17) (geo,9.5) (mark,-6)};
        \end{groupplot}

\node[
    draw=black, 
    fill=white, 
    font=\scriptsize, 
    inner sep=3pt, 
    align=center, 
    yshift=-20pt, 
] at ($(group c1r3.south)!0.5!(group c3r3.south)$) 
{ \tikz \draw[fill=white, draw=black, pattern=staggereddots, pattern color=black] (0,0) rectangle (0.3,0.15); \ validation \quad
    \tikz \draw[fill=mplred!20, draw=black, pattern=north east lines, pattern color=black] (0,0) rectangle (0.3,0.15); \ test
    \quad
    \tikz \draw[fill=mplred] (0,0) rectangle (0.3,0.15); \ rule-based
    \quad
    \tikz \draw[fill=mplblue] (0,0) rectangle (0.3,0.15); \ ArgoTweak};
    \end{tikzpicture}

    \caption{When training on rule-based priors (red), the sim2real gap (third row) is substantially larger than when trained on ArgoTweak (yellow).}
    \label{fig:sim2realgap}
\end{figure}
While the first experiment demonstrated why change-aware modeling is necessary in the first place, we now show that even change-aware models fail to generalize when trained on scripted priors -- highlighting the need for ArgoTweak’s realism to close the sim2real gap.
In this experiment, we train two versions of our network with primary and secondary change assessment heads. Given that noise-based synthetic priors do not align with meaningful change categories (see \cref{sec:syntheticpriorcomparison}), we conduct a comparative analysis between training on ArgoTweak and on rule-based priors~\cite{tbv}. Since both datasets describe realistic changes, we apply atomic change reasoning.
\begin{table*}[h]
    \centering
    \begin{tabular}{l|cccccc|c|c|ccccc|c}
        \toprule
         \multirow{2}{*}{\centering Changes} & \multicolumn{6}{c|}{$\text{mAP}_c$} & \multirow{2}{*}{\centering mAP} & \multirow{2}{*}{\centering $\text{mAPC}$} & \multicolumn{5}{c|}{$\text{mAcc}_c$} & \multirow{2}{*}{\centering mACC}\\
          & $\neg c$ & $c$ & ins & del & geo & mark & & & $c$ & ins & del & geo & mark &\\
        \midrule
        (1)\ none  & -- &  -- & -- &  -- &   -- & -- & 77.7 & -- &  -- & -- &  -- & -- &  -- & -- \\
        (2)\ $c/\neg c$\textsuperscript{\dag}  & 79.0 & 14.7 & -- &  -- &   -- & -- & 77.5 & 46.9 & 66.5 & -- &  -- & -- &  -- & 66.5\\
        (3)\ $c/\neg c$ & 79.0 & 10.4 & -- &  -- & -- &  -- & 77.7 & 44.7 & 70.5 & -- &  -- & -- &  -- &70.5\\
        (4)\ full\textsuperscript{\dag}  & 80.5 & 19.7 & 9.4 & 15.7  & --  & 5.1\textsuperscript{\dag} & 79.4 & 10.1\textsuperscript{\dag} & 71.0 & 67.0 & 60.5 & -- & 65.5\textsuperscript{\dag} & 64.3\\
        (5)\ full\textsuperscript{\ddag}  & 77.4 & 16.0 & 9.0 & 16.8 & \multicolumn{2}{c|}{5.1\textsuperscript{\ddag}}   & 78.3 & 10.3\textsuperscript{\ddag} & 71.5 & 67.0 & 60.5 & \multicolumn{2}{c|}{65.5\textsuperscript{\ddag}} & 64.3\\
        \midrule
        (6)\ full & 78.5 & 14.0 & 7.6 & 16.3  & 4.6 & 6.3 & 78.8 & 8.7 & 71.5 & 68.5 & 61.0 & 68.0 & 61.0 & 64.6\\
        \bottomrule
    \end{tabular}
    \caption{Performance comparison for ArgoTweak-trained models with different levels of annotation detail: no change annotation (\textit{none}), binary change annotation ($c/\neg c$) and atomic change annotation (\textit{full}). \textsuperscript{\dag} models trained on ArgoTweak without annotation of geometry changes. \textsuperscript{\ddag} primary change detection head only. Notably, mAPC and mACC are computed over different sets of $\mathcal{C}$, with $\mathcal{C}=\{c, \neg c\}$ for models 2 and 3, and $\mathcal{C}=\{\text{ins},\text{del},\text{geo\textsuperscript{(\dag)}},\text{mark}\}$ for models 4-6.}
    \label{tab:ablation_tab}
\end{table*}

To identify sim2real  gaps, we evaluate both networks on the real-world test split, as well as on the ArgoTweak validation set and the rule-based validation set, respectively. We compute the absolute difference in $\text{mAcc}_c$ for $c \in \{\text{insertion, deletion, geometry change, mark change}\}$. The results, presented in \cref{fig:sim2realgap}, indicate that the model trained on rule-based priors exhibits significantly larger sim2real gaps across all metrics. For the combined metric of $\text{mACC}_c$, we observe $\Delta\text{mACC}=-36.0$ for the model trained on a rule-based prior, whereas our ArgoTweak dataset reduces this gap by more than a factor of 10 to $\Delta\text{mACC}=-3.5$. Additional results computed on our fine metric mAPC are shown in \ref{sec:moreresults}.

\subsection{Ablation studies}
\begin{figure}
    \centering
    \setlength{\tabcolsep}{0pt}

    \begin{minipage}{0.14\textwidth}
        \begin{subfigure}{1\textwidth}
        \centering
        \fboxsep=0pt
        {\rotatebox{-90}{\includegraphics[width=\textwidth]{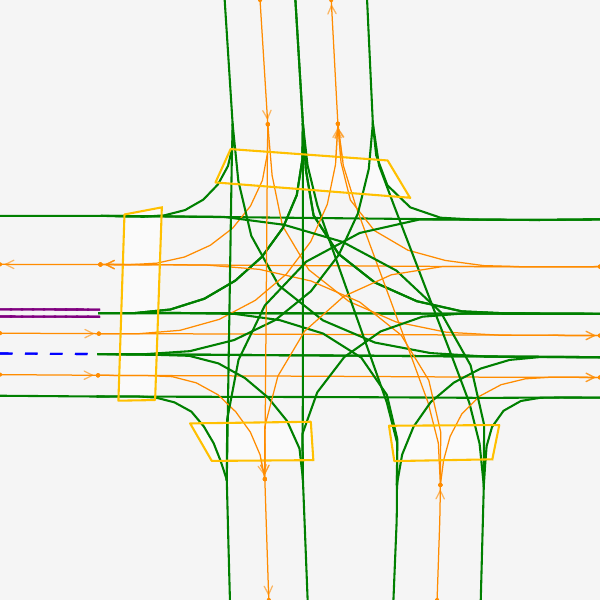}}}
        \caption{map prior}
        
    \end{subfigure}
    \end{minipage}
    \hfill
    \begin{minipage}{0.30\textwidth}
        \centering
        \begin{minipage}{0.48\textwidth}
            \begin{subfigure}{1\textwidth}
        \centering
        \fboxsep=0pt
        {\rotatebox{-90}{\includegraphics[width=\textwidth]{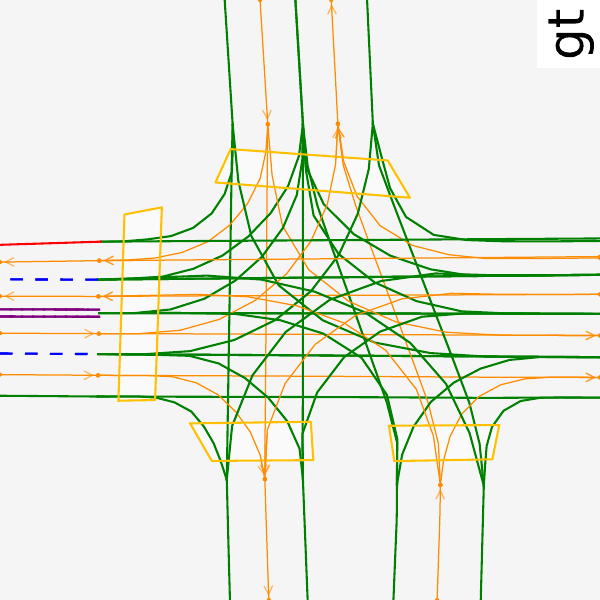}}}
        
    \end{subfigure}
        \end{minipage}
        \begin{minipage}{0.48\textwidth}
            \centering
            \rotatebox{-90}{\adjustbox{padding=0.5pt}{\includegraphics[width=\textwidth]{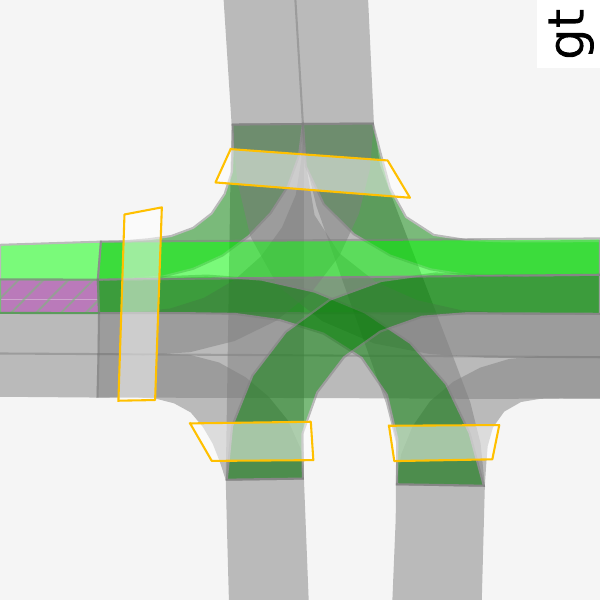}}}
        \end{minipage}

        \begin{minipage}{0.48\textwidth}
             \begin{subfigure}{1\textwidth}
        \centering
        \fboxsep=0pt
        {\rotatebox{-90}{\includegraphics[width=\textwidth]{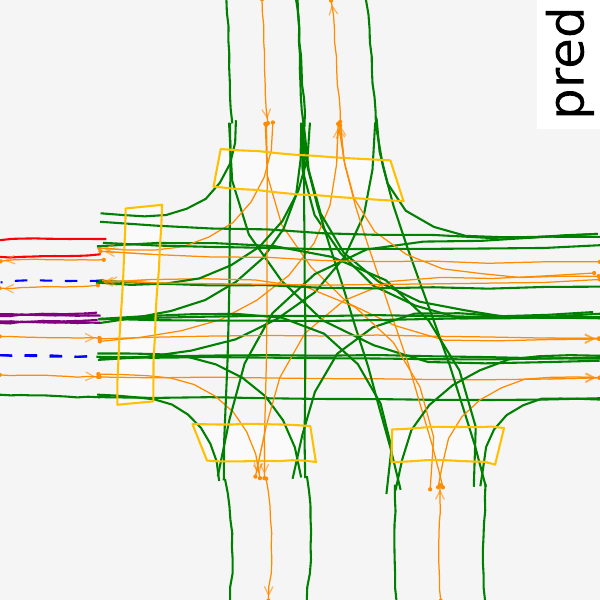}}}
        \caption{map update}
        
    \end{subfigure}
        \end{minipage}
        \begin{minipage}{0.48\textwidth}
            \begin{subfigure}{1\textwidth}
        \centering
        \fboxsep=0pt
        {\rotatebox{-90}{\includegraphics[width=\textwidth]{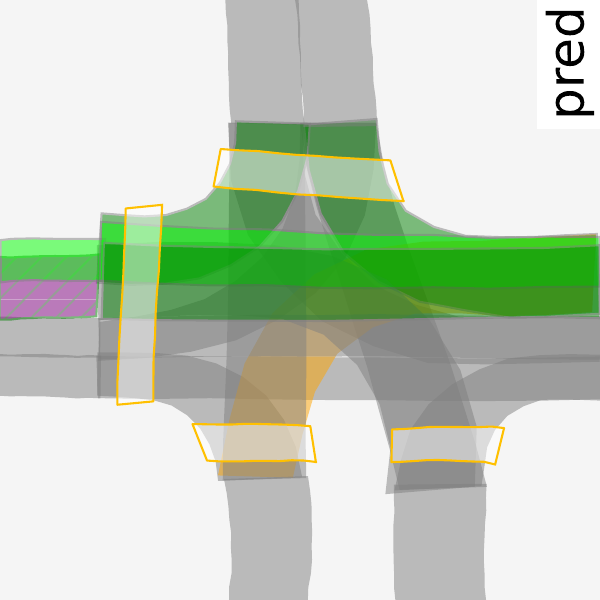}}}
        
            \caption{change assessment}
            \end{subfigure}
        \end{minipage}
    \end{minipage}

    \caption{Example of our ArgoTweak-trained model. For change assessment, purple denotes marking changes, light green insertions, dark green geometry edits. Yellow elements were classified as "Other", but no secondary head was triggered.}
    \label{fig:example}
\end{figure}
We investigate how varying levels of annotation detail affect performance. 
Notably, models trained under different annotation granularities cannot be directly compared: As detailed in \cref{sec:refinements}, search boundaries for element merging depend on the change status, meaning annotation granularity alters lane segment partitions.

For models trained without change annotations (\cref{tab:ablation_tab},1), the only available performance indicator is mAP (\cf~\cref{sec:syntheticpriorcomparison}). This value is higher for networks trained \textit{with} full change-aware approaches (\cref{tab:ablation_tab}, 4-6). Within the constraints imposed by different merging conditions, this suggests that change annotation not only facilitates detailed evaluation but also aids the network during training.

Next, we compare two networks trained with binary change labels and the primary change assessment head to distinguish between changed and unchanged elements (\cref{tab:ablation_tab}, 2 and 3). Here, model 2 excludes geometric changes from the annotations. This exclusion is motivated by the observation that current map generation methods often lack the geometric precision necessary to reliably distinguish subtle road shape modifications from noise. Configuration 2 achieves comparable mAP and $\text{mAP}_{\neg c}$ but outperforms model 3 in terms of mAPC and $\text{mAP}_c$. This suggests that omitting geometric changes enables the network to better predict the shape of changed elements. Interestingly, the model’s mean accuracy (mACC) decreases, indicating that overall change detection is negatively impacted when deviations between prior and predicted maps are not annotated.

Finally, we evaluate models trained with fine-grained atomic change annotations (\cref{tab:ablation_tab}, 4–6 and \cref{fig:example}). 
In this experiment, we remove the secondary change assessment head (model 5), treating marking and geometric changes as a single category. While differences in map element merging prevent direct numerical comparisons, we do not observe significant performance degradation across key metrics. This suggests that our bijective mapping framework effectively captures the diversity of real-world map changes, ensuring that the annotation strategy does not introduce confusion, even with the full set of change categories.

\section{Conclusion}
We introduced  ArgoTweak, the first dataset to pair realistic map priors with current sensor data and up-to-date ground-truth maps. With our bijective mapping framework, we structured map updates into an explainable process, enabling fine-grained annotations that distinguish real-world changes from stable elements. Through extensive experiments, we demonstrated that models trained with ArgoTweak significantly reduce the sim2real gap, while our fine-grained metrics, mAPC and mACC, unlocked deep insights into the balance between map stability and adaptability. By setting a new benchmark for self-updating HD maps, ArgoTweak advances scalable, explainable, and continuously improving mapping solutions. 
\newpage
\section*{Acknowledgements}
The research work was funded by the Swedish Foundation for Strategic Research (SSF) under the project DeltaMap (ID22-0045). This work was partially supported by the Wallenberg AI, Autonomous Systems and Software Program
(WASP) funded by the Knut and Alice Wallenberg Foundation. We thank Mohammad Nazari for his feedback on this work and the technical support.
{
    \small
    \bibliographystyle{ieeenat_fullname}
    \bibliography{main}
}
\newpage
\definecolor{mplblue}{rgb}{1, 0.6470588235, 0}
\definecolor{mplred}{rgb}{1,0,0}
\maketitlesupplementary
\section{Bijective mapping} \label{sec:bijectivedetails}
This appendix provides additional details and insights into the \textbf{ArgoTweak Dataset}, extending the main paper’s discussion on its design and application. As introduced in \cref{sec:dataset}, the dataset is motivated by the need for \textbf{systematic, high-fidelity change annotations} in HD map updates. Current datasets used for prior-aided mapping lack structured change annotations, limiting model refinement and evaluation. To address this, we introduce the concept of \textbf{bijective change mapping}, a framework that ensures a \textbf{one-to-one correspondence} between high-level structural updates and their corresponding atomic changes at the element level.

\subsection{Key definitions}
As detailed in \cref{sec:bijectivityrecs}, our framework operates on different map levels. Below, we provide detailed descriptions with examples:
\begin{itemize}
    \item \textbf{Atomic change $a_i$}: The smallest indivisible edit applied to individual map element (e.g., attribute modifications, vertex adjustments). These edits involve exactly one map element, without influencing the neighboring structures.
    \item \textbf{Structural updates $y_i$}: Large-scale updates that affect multiple elements, such as adding lanes, modifying intersections, or adjusting entire road layouts. These modifications are explainable through one or multiple atomic changes on one or multiple map elements.
    \item \textbf{Set of atomic changes $x_i$}: A specific permutationally invariant set of atomic changes, which explains a structural update through applying one or several element-wise atomic edits.
    \item \textbf{Macro-modification $\hat{y}_i$}: Large-scale modifications that form a subset of $\hat{\mathbf{Y}}$, whose combinations approximate $\textbf{Y}$. These modifications affect shape, appearance, function of map parts, as well as their underlying lane graph and the total lane number in a local patch.
    \item \textbf{Bijective change mapping $f$}: A novel framework that establishes a direct, one-to-one mapping between structural updates and atomic changes, ensuring consistency and traceability in prior-aided mapping.
\end{itemize}

\subsection{Desiderata for explanation of modifications through atomic change categories}
\label{subsec:desiderata}
The goal is to ensure a bijective mapping between structural updates on the HD map and their explanation through atomic changes. 
We define the desiderata for this mapping in \cref{tab:desiderata}.

\begin{table*}[th!]
\centering
\renewcommand{\arraystretch}{1.2}
\setlength{\tabcolsep}{8pt}
\begin{tabular}{p{0.5cm}|p{2.1cm}|p{7.7cm}|p{4.5cm}}
\toprule
ID & Statement & Requirement & Benefit \\
\midrule
$\text{A}_1$ & Uniqueness and distinctiveness & Ensure that each atomic change produces a unique, interpretable outcome in the map. No two types of changes should lead to the same map state. Every modification should have a distinct representation. & Prevents ambiguity in training map updating algorithms that rely on clear distinctions between map states. \\
\midrule
$\text{A}_2$& Non-redundancy & Avoid overlapping or redundant categories that could confuse the classification process or lead to interpretive flexibility. Each category should capture a unique aspect of the modification, avoiding overlap with other categories. & Reduces confusion for both human operators and automated systems interpreting the changes through streamlined categories. \\
\midrule
B & Relevance to functionality & Reflect changes that significantly impact how the map would be used by autonomous systems, particularly for navigation and safety. Categories should capture changes that affect traffic flow, lane usage, directional changes, or functionality (e.g., converting a vehicle lane to a bike lane). & Ensures that the map's usability and safety are prioritized, with critical modifications being flagged and interpreted appropriately. \\
\midrule
C & Hierarchical change tracking & Facilitate tracking of both high-level and detailed modifications, allowing for a hierarchical understanding of changes. Changes should be categorized at both coarse and fine levels. For example, a high-level category might indicate “structural change” while a subcategory specifies “lane deletion.” & Enhances interpretability for complex modifications and allows for selective filtering of changes based on granularity. \\
\midrule
D & Modular change tracking & Maintain accuracy and clarity even in complex scenarios, such as multi-lane adjustments or simultaneous functional and geometric changes, by a modular approach to complex changes. The framework must handle composite changes without ambiguity.  
& Supports classification of multi-layered modifications, preventing misinterpretation when an element undergoes several modifications. \\
\bottomrule
\end{tabular}
\caption{Mapping requirements and benefits.}
\label{tab:desiderata}
\end{table*}
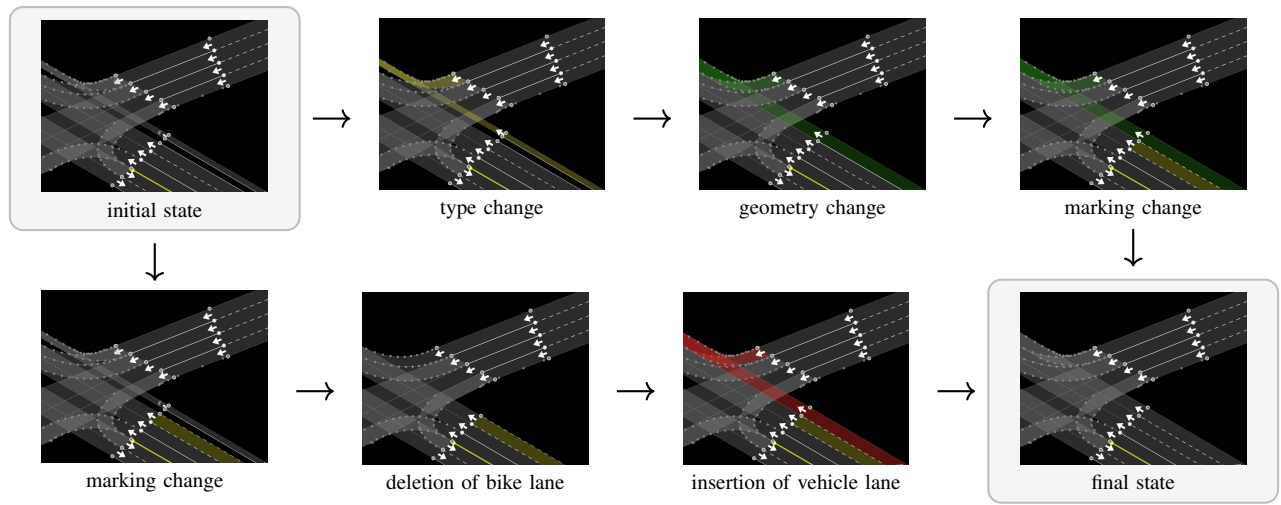
\begin{figure*}[h!] 
    \centering
    \begin{tikzpicture}[
        node distance=0.5cm and 0.5cm,
        every node/.style={text width=3.5cm, align=center, font=\footnotesize},
        every path/.style={draw, ->, thick}
    ]
    
    \node (start) at (0,0)[draw=lightgray, fill={rgb,1:red,0.96;green,0.96;blue,0.96}, thick, inner sep=5pt, outer sep=5pt, rounded corners=5pt]{\includegraphics[width=3cm]{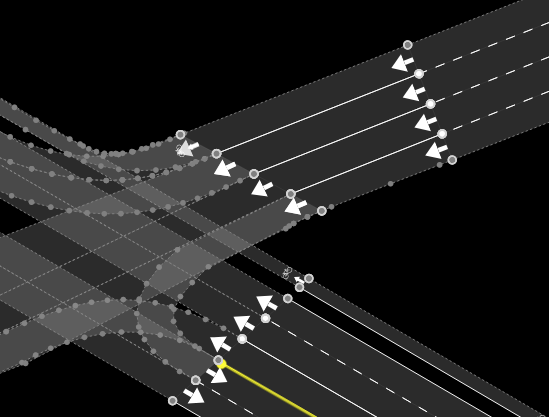} \\ initial state};
    
    \node (mod4a) [right=of start] {\includegraphics[width=3cm]{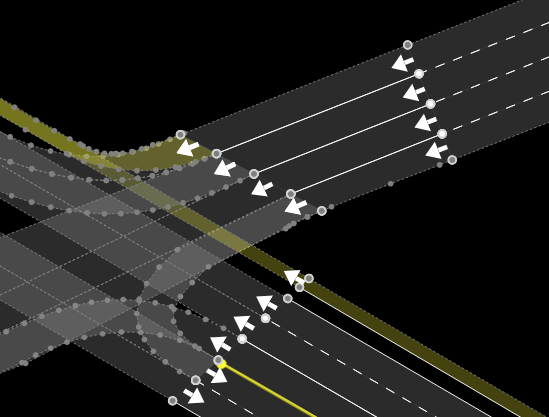} \\ type change};
    \node (mod4b) [right=of mod4a] {\includegraphics[width=3cm]{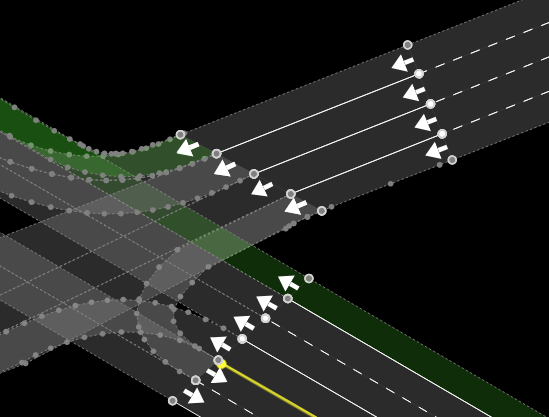} \\ geometry change};
    \node (mod4c) [right=of mod4b] {\includegraphics[width=3cm]{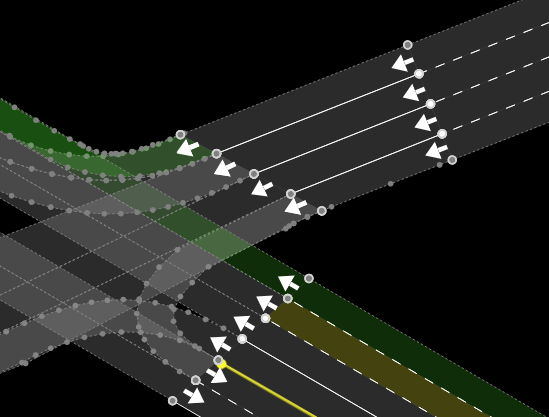} \\ marking change};
    
    \node (mod5a) [below=of start] {\includegraphics[width=3cm]{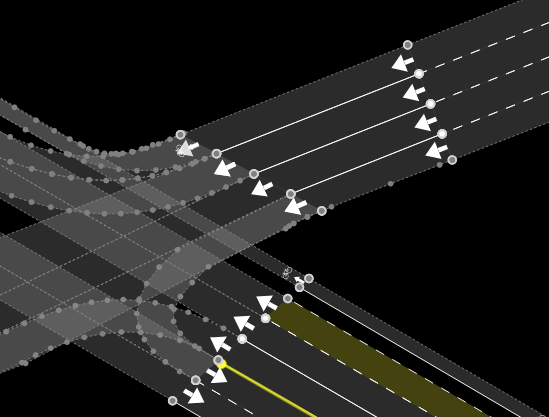} \\ marking change};
   \node (mod5b) [right=of mod5a] {\includegraphics[width=3cm]{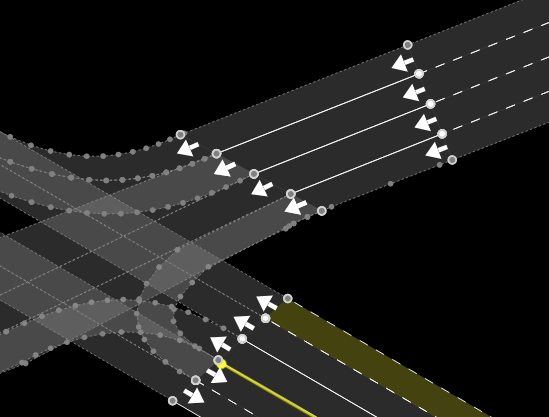} \\ deletion of bike lane};
   \node (mod5c) [right=of mod5b] {\includegraphics[width=3cm]{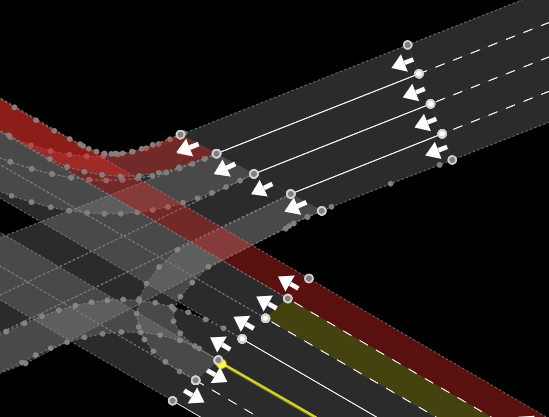} \\ insertion of vehicle lane};
   
    \node (end) [below=of mod4c,draw=lightgray, fill={rgb,1:red,0.96;green,0.96;blue,0.96}, thick, inner sep=5pt, outer sep=5pt, rounded corners=5pt] {\includegraphics[width=3cm]{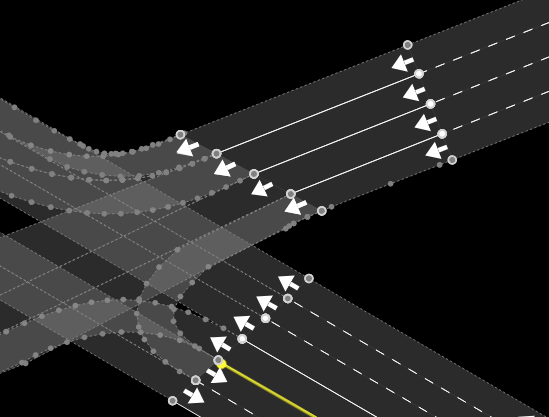} \\ final state};

    \path (start) edge (mod4a);
    \path (start) edge (mod5a);
   
    \path (mod4a) edge (mod4b);
    \path (mod4b) edge (mod4c);
    \path (mod4c) edge (end);
    \path (mod5a) edge (mod5b);
    \path (mod5b) edge (mod5c);
    \path (mod5c) edge (end);

    \end{tikzpicture}
    \caption{Motivation for bijectivity (\cref{sec:motbij}). Green elements have been geometry-edited, red elements are insertions and yellow elements underwent type or marking-related changes.}\label{fig:atomicity_fig}
\end{figure*}
\subsection{Map element properties}
The elementary units of our approach are lane segments and pedestrian crossings. Notably, in our model backbone \cite{LaneSegNet}, their representation is unified by treating pedestrian crossings as lane segments oriented perpendicular to the driving direction. A lane segment consists of a left border, a right border, and a set of properties. These properties are detailed in \cref{tab:lane_properties}, following and complementing the Argoverse 2 map format \cite{Argoverse2}.
\subsection{Motivation for bijectivity}\label{sec:motbij}
From a map annotation perspective, we want to retrieve the richest possible set of change annotations, in the sense that we want to avoid insertions and deletions where possible. This is because an element replacement, \ie deleting an existing map element, and inserting a new one from scratch, "wipes out" potential correspondences between the prior and the ground-truth map. This could limit the ability to leverage structured priors effectively. 
\begin{table*}[h!]
    \centering
    \begin{tabular}{l|c|c|p{11cm}}
        \toprule
        Property & ls & pc & Description \\
        \midrule
        \texttt{id} &\checkmark&\checkmark&  A unique identifier for this lane segment or pedestrian crossing. \\
        \midrule
        \texttt{is\_intersection} &\checkmark& & True if the lane segment is part of an intersection. \\
        \midrule
        \texttt{lane\_type} &\checkmark&  & Specifies the type of lane (vehicle, bike, bus). \\
        \midrule
        \texttt{left\_lane\_boundary} &\checkmark&\checkmark  & List of points defining the left boundary of the lane. \\
        \midrule
        \texttt{right\_lane\_boundary} &\checkmark& \checkmark & List of points defining the right boundary of the lane. \\
        \midrule
        \texttt{centerline} &\checkmark&\checkmark& List of points representing the geometric center of the lane or pedestrian crossing. \\
        \midrule
        \texttt{left\_lane\_mark\_type} &\checkmark&  & Type of marking on the left lane segment boundary with mark type solid/dashed/double-solid/double-dashed/dash-solid/solid-dash/none/unknown and mark color white/yellow/blue/non-visible. \\
        \midrule
        \texttt{right\_lane\_mark\_type} &\checkmark&  & Type of marking on the right lane segment boundary with mark type solid/dashed/double-solid/double-dashed/dash-solid/solid-dash/none/unknown and mark color white/yellow/blue/non-visible. \\
        \midrule
        \texttt{successors} &\checkmark&& Array of IDs for lane segments that follow the segment. \\
        \midrule
        \texttt{predecessors} &\checkmark&&  Array of IDs for lane segments that precede the segment. \\
        \midrule
        \texttt{right\_neighbor\_id} &\checkmark&  & ID of the lane immediately to the right, if it exists. \\
        \midrule
        \texttt{left\_neighbor\_id} &\checkmark&  & ID of the lane immediately to the left, if it exists. \\
        \midrule
        \texttt{is\_modified} &\checkmark&\checkmark& Indicates whether the segment has been modified compared to the base map. \\
        \midrule
        \texttt{change\_hist} &\checkmark&\checkmark& A list of changes applied to the lane segment or pedestrian crossing over time. \\
        \bottomrule
    \end{tabular}
    \caption{Lane segment (ls) and pedestrian crossing (pc) object definitions.}
    \label{tab:lane_properties}
\end{table*}

\begin{figure*}[t!]
    \centering
    \begin{tikzpicture}[
        node distance=1.5cm and 0.5cm,
        every node/.style={text width=3.5cm, align=center, font=\footnotesize},
        every path/.style={draw, ->, thick}
    ]

    \node (start) at (0,0) [draw=lightgray, fill={rgb,1:red,0.96;green,0.96;blue,0.96}, thick, inner sep=5pt, outer sep=5pt, rounded corners=5pt]{\includegraphics[width=3cm]{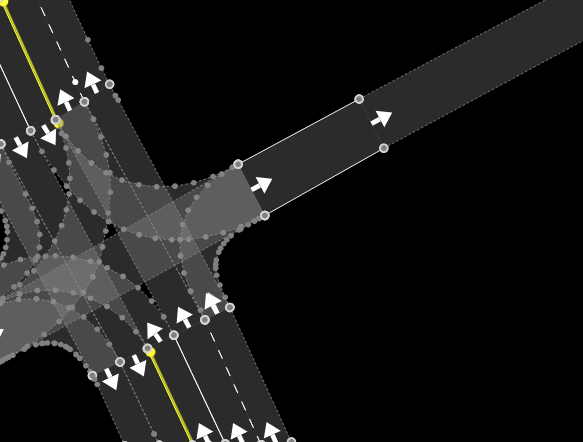} \\ initial state};

    \node (mod1a) [right=of start,  yshift=3cm] {\includegraphics[width=3cm]{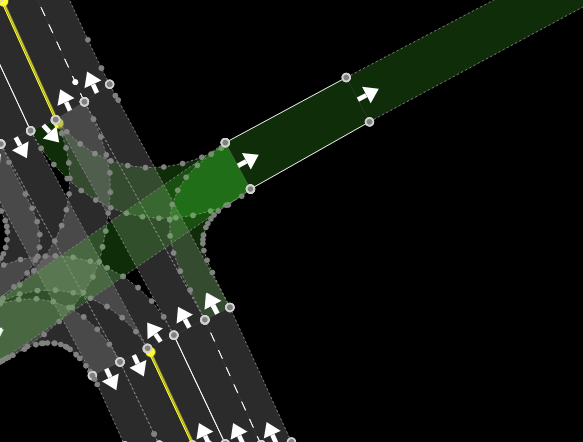} \\ geometry change};
    \node (mod1b) [right=of mod1a] {\includegraphics[width=3cm]{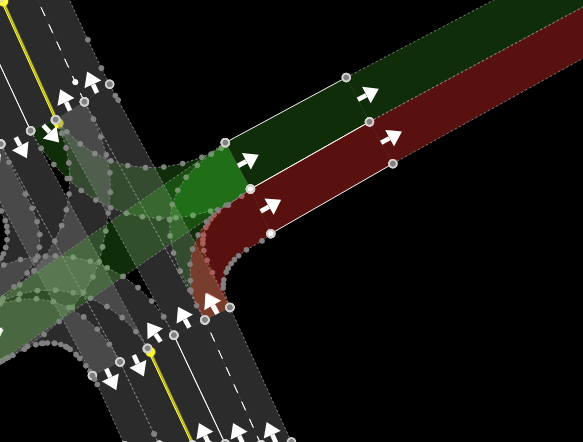} \\ insertion of right lane};
    
    \node (mod2a) [right=of start] {\includegraphics[width=3cm]{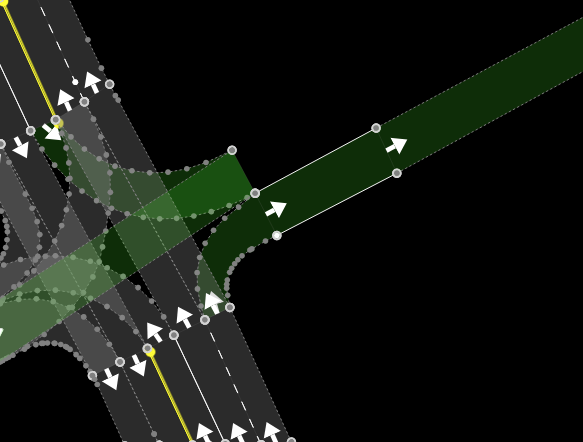} \\ geometry change};
   \node (mod2b) [right=of mod2a] {\includegraphics[width=3cm]{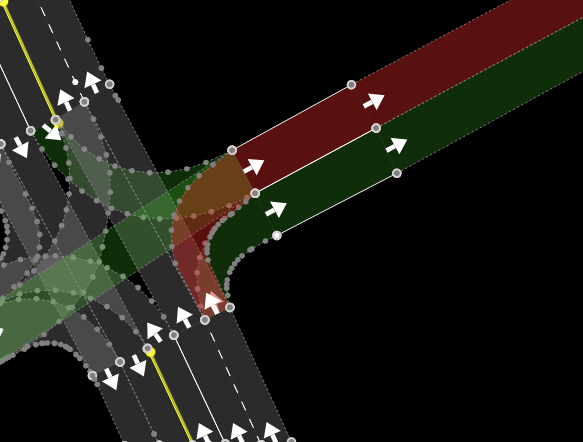} \\ insertion of left lane};
   
    \node (mod3a) [right=of start, yshift=-3cm] {\includegraphics[width=3cm]{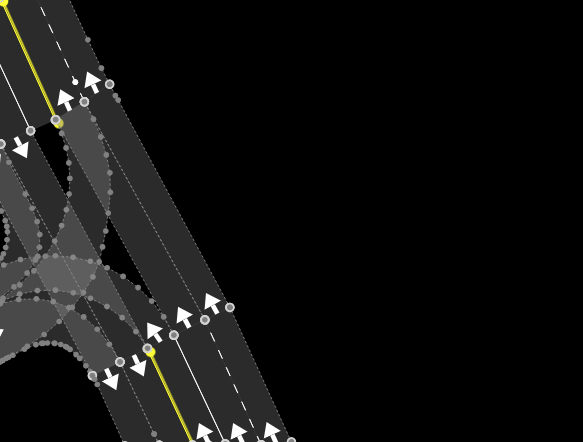} \\ deletions};
   \node (mod3b) [right=of mod3a] {\includegraphics[width=3cm]{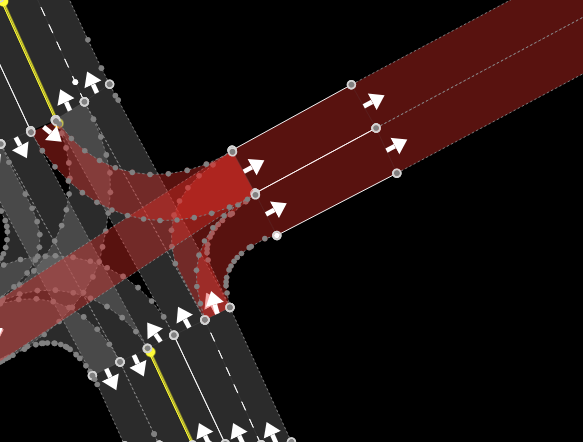} \\ replacements};

    \node (end) [right=of mod1b,  yshift=-3cm, draw=lightgray, fill={rgb,1:red,0.96;green,0.96;blue,0.96}, thick, inner sep=5pt, outer sep=5pt, rounded corners=5pt] {\includegraphics[width=3cm]{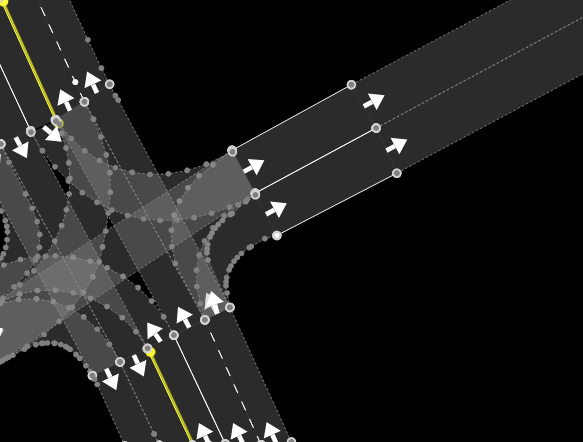} \\ final state};

    \path (start) edge (mod1a);
    \path (start) edge (mod2a);
    \path (start) edge (mod3a);
   
    \path (mod1a) edge (mod1b);
    \path (mod1b) edge (end);
    \path (mod2a) edge (mod2b);
    \path (mod2b) edge (end);
    \path (mod3a) edge (mod3b);
    \path (mod3b) edge (end);

    \end{tikzpicture}
    \caption{In the above figure, green elements have been geometry-edited, whereas red elements are insertions. The Right-Handside-Rule in \cref{sec:rhsr} defines the top path as the only viable option, ruling out the central path. The bottom path is prohibited by our bijective mapping framework, because the underlying road graph does not justify element replacements.}\label{fig:righthandsiderule}
\end{figure*}
In \cref{fig:atomicity_fig}, we illustrate such an example. The structural modification on the shown map patch describes a bike lane turning into an additional vehicle lane. In the upper modification path, we first change the lane type from "BIKE" to "VEHICLE". Next, we modify the geometry of the segments involved. Finally, we alter the lane line marking type for one of the original vehicle lanes. With this approach (which is the one supported by our bijective mapping framework), we maintain correspondence between elements in the prior and their updated versions in the ground truth map. Contrarily, in the lower modification path, we completely remove the bike lane first, and insert a new lane, although the underlying road graph is unchanged. Hence, this is prohibited by our framework.

\subsection{The Right-Handside-Rule}\label{sec:rhsr}
Since atomic changes define which edits to apply to a map element, but not necessarily their global location and direction, we standardize insertions and deletions to begin in the driving direction right in cases where it is not immediately clear which element is the inserted one. An example motivating this so-called Right-Handside-Rule can be seen in \cref{fig:righthandsiderule}: The top, central and bottom modification path lead to the same final map state. By applying our bijective mapping, we can discard the bottom path, because deleting the existing structures and replacing them is not justified given the underlying road graph: In both prior and ground-truth-map, the road graph includes at least one lane travelling to the right. However, both central and top path are compatible with our framework. To introduce a uniform map annotation strategy, we apply the Right-Handside-Rule, leaving the top path as the only solution compatible with our framework.

\begin{figure*}[h]
    \centering
    
    \begin{tikzpicture}[every node/.style={text width=3.5cm, align=center, font=\footnotesize}]

        \draw[draw=lightgray, fill={rgb,1:red,0.96;green,0.96;blue,0.96}, thick, inner sep=5pt, outer sep=5pt, rounded corners=5pt] (-2, 1.7) rectangle (2, -4.7); 

        \node[inner sep=0pt] (img1) at (0, 0) {\includegraphics[width=3cm]{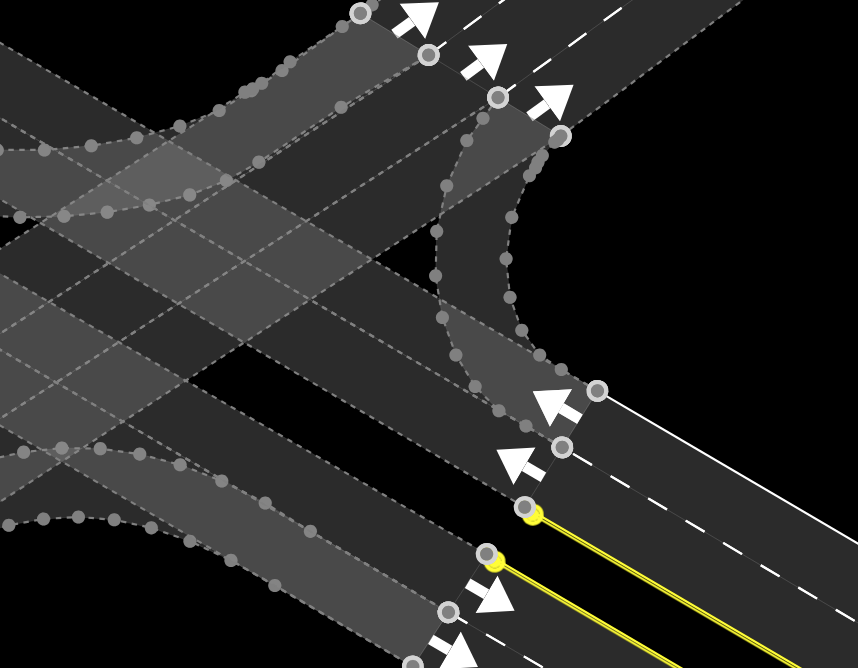}};
        \node[inner sep=0pt] (img2) at (4.5, 0) {\includegraphics[width=3cm]{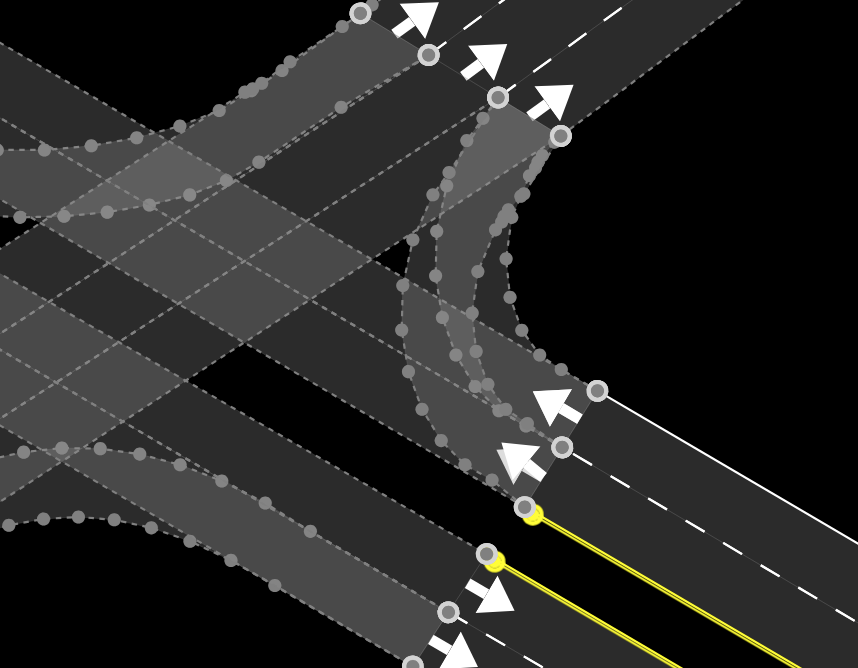}};
        \node[inner sep=0pt] (img3) at (8.25, 0) {\includegraphics[width=3cm]{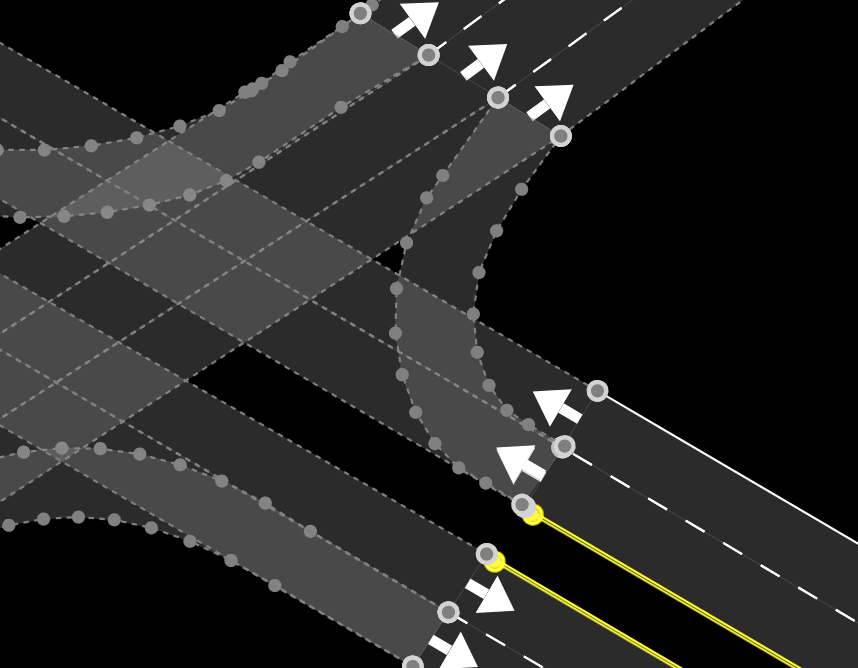}};
        \node[inner sep=0pt] (img4) at (12, 0) {\includegraphics[width=3cm]{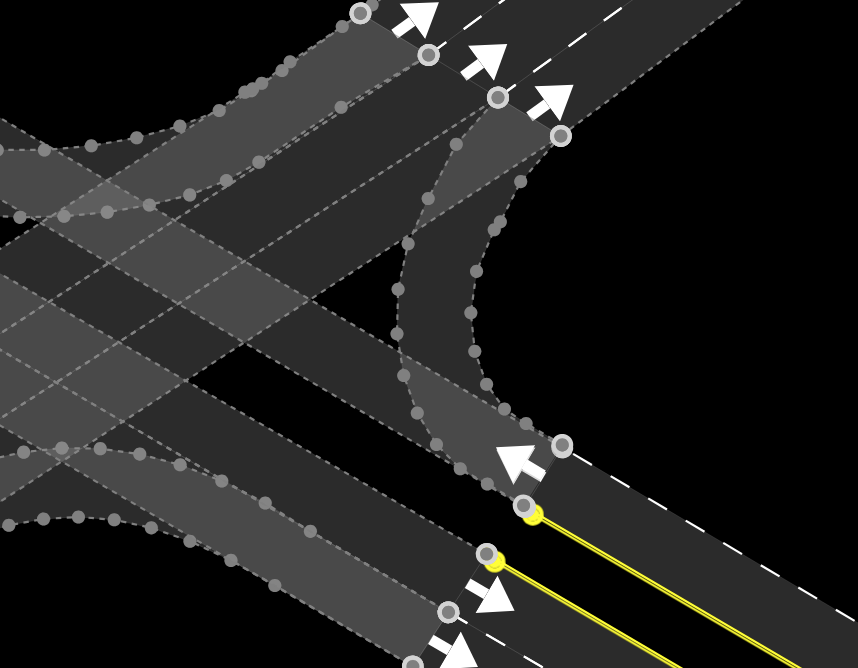}};

        \node[inner sep=0pt] (img5) at (0, -3) {\includegraphics[width=3cm]{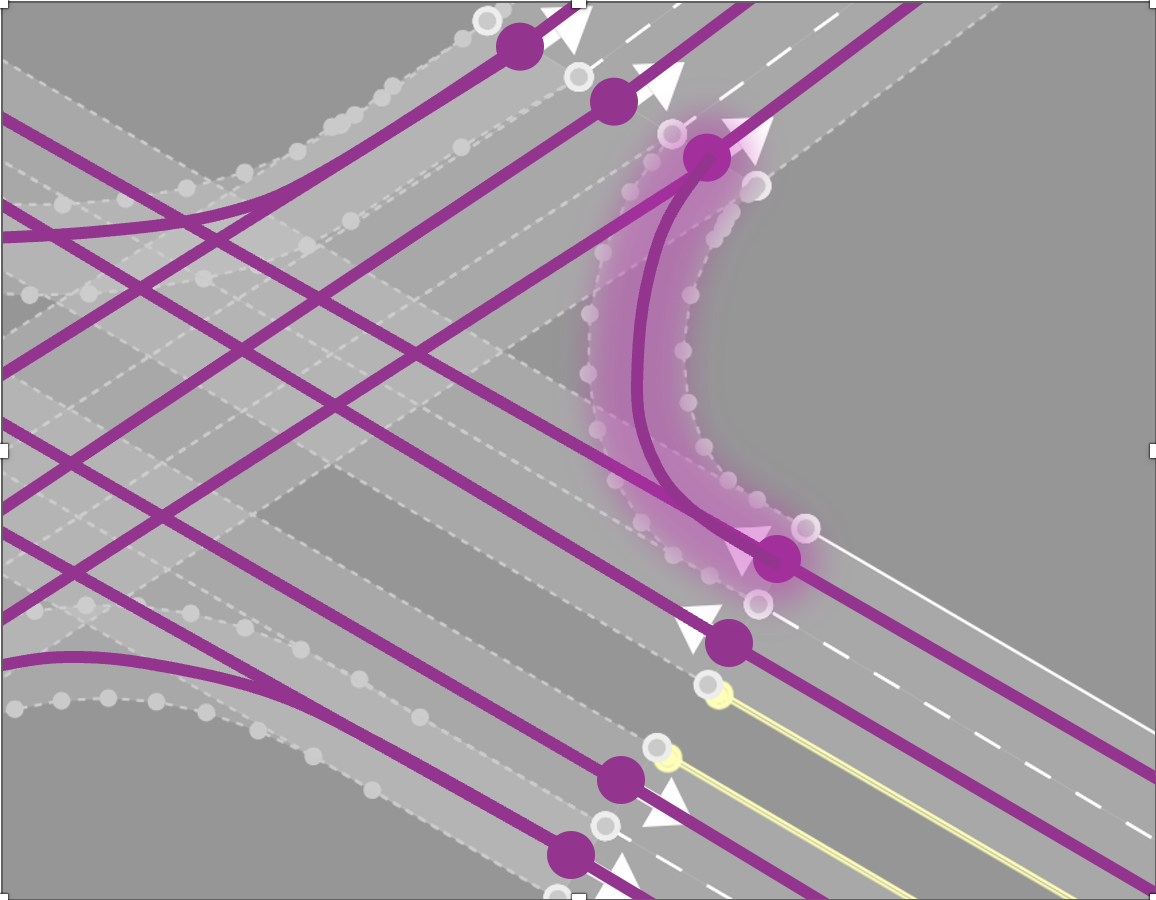}\\ ground truth};
        \node[inner sep=0pt] (img6) at (4.5, -3) {\includegraphics[width=3cm]{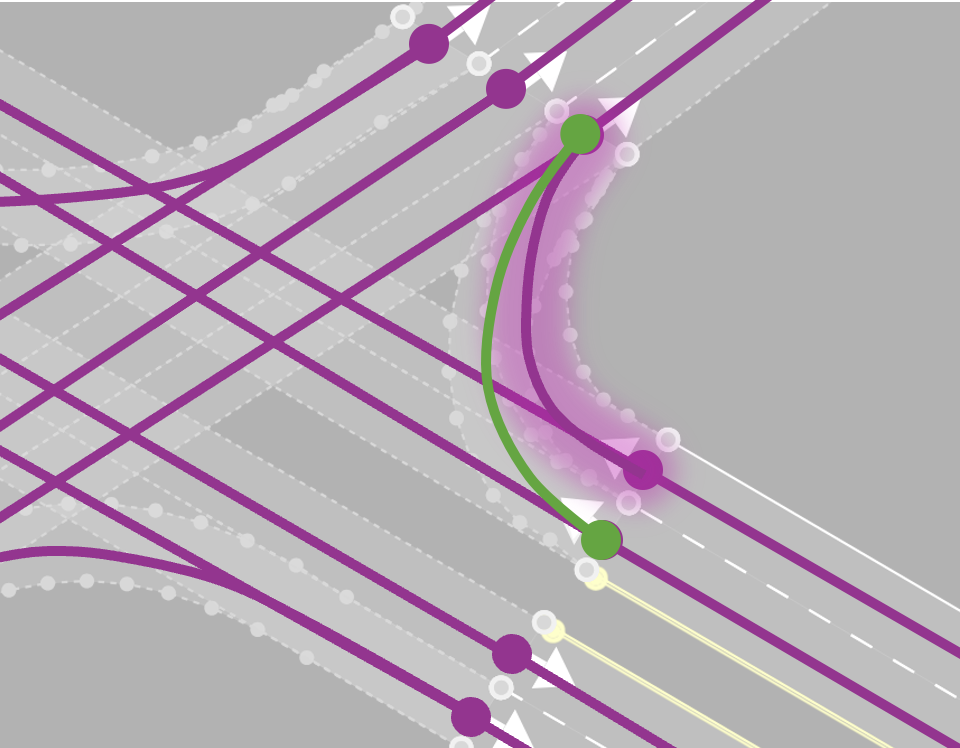}\\(1) insertion};
        \node[inner sep=0pt] (img7) at (8.25, -3) {\includegraphics[width=3cm]{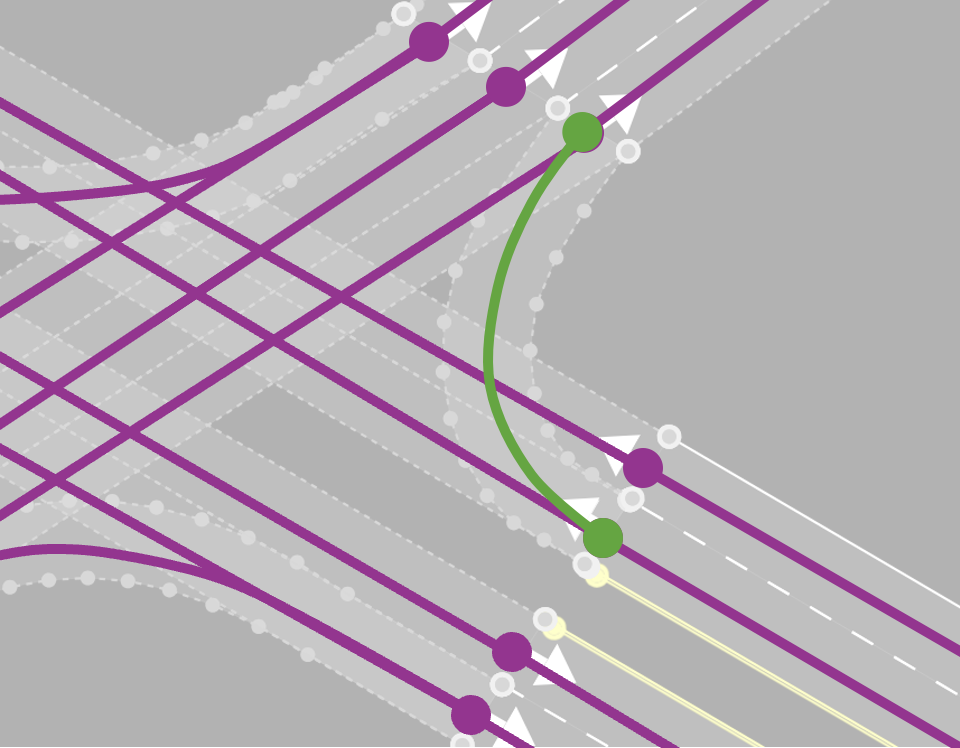}\\(2) re-routing};
        \node[inner sep=0pt] (img8) at (12, -3) {\includegraphics[width=3cm]{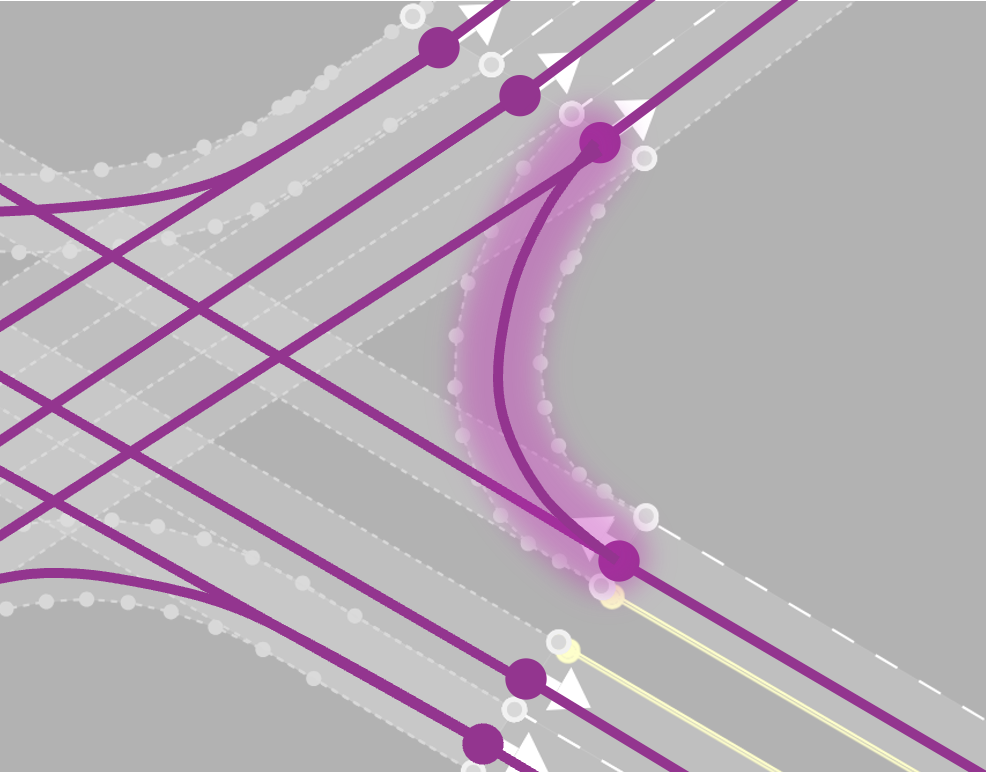}\\(3) deletion};

    \end{tikzpicture}
    
    \caption{Illustration of different road-graph altering changes. The highlighted pink lane segment defines a right-turn from the outmost lane. Green connections indicate inserted element. A detailed discussion can be found in \cref{sec:topofunc}.}
    \label{fig:topofunc}
\end{figure*}
\subsection{Function-preserving changes in intersections} \label{sec:topofunc}
 We note that within intersections, the road graph is often altered, leading to a large number of deletion+insertion pairs. To prevent overly complex annotations within intersections, we allow element re-routing as a special type of geometry modification. The condition for this special annotation category is that the topological function of the element defined on the road-graph remains unchanged (\eg, turning left from the outermost lane, continuing straight from the center lane). 

The concept of these function-preserving road-graph changes is best illustrated through the example in \cref{fig:topofunc}, which presents three different scenarios and how they are annotated in our framework while allowing function-preserving changes in intersections.

In \cref{fig:topofunc}, (1), a new right-turn lane is inserted, changing the total number of lanes in the map. This case is classified as an insertion, as the function of the existing lanes remains unchanged, but the number of lanes in the global road increased, which is accounted for within our framework. Note that the insertion is not constrained by the Right-Handside Rule, as it is immediately clear which is the inserted element, while all existing elements remain unchanged.

In \cref{fig:topofunc}, (2), the right-turn lane is no longer accessible from the outermost lane, but re-routed to the inner lane. While the number of lanes remains unchanged, this re-routing fundamentally alters the functional aspect of the road. Here, the modification is represented by a replacement, \ie, an insertion+deletion pair, as the lane connectivity is altered in a way that impacts traffic behavior.

In \cref{fig:topofunc}, (3), the right-turn option is still removed from the outer lane, but an additional change occurs: the number of lanes in the driving direction is reduced from two to one. In this case, the lane graph structure changes, while the highlighted lane segment in question maintains its original topological function on the new road graph: It allows for right-turns from the right-most lane. This scenario demonstrates a function-preserving road-graph change, where the structure is adjusted but the intended driving options remain clear. Hence, we annotate the segment in question with a geometry change. 
 In our dataset, we mark these elements as special geometry edits, allowing for easily converting them back into insertion-deletion pairs if needed.
\subsection{Annotations beyond atomic changes}
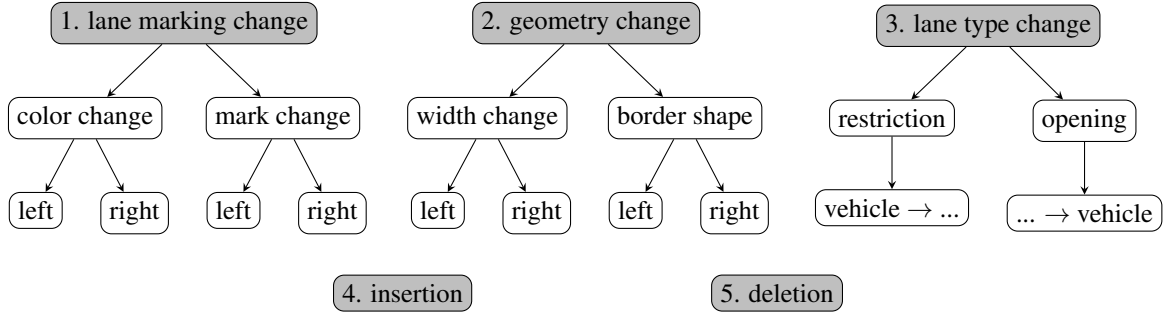
\begin{figure*}[h]
    \centering
    \begin{minipage}{0.3\textwidth}
    \centering
    \begin{forest}
    for tree={
        grow=south,
        edge={-stealth},
        draw,
        rounded corners,
        node options={align=center},
        s sep=5mm,
        l sep=7mm
    }
    [1. lane marking change, fill=lightgray
      [color change
        [left]
        [right]
      ]
      [mark change
        [left]
        [right]
      ]
    ]
    \end{forest}
    \end{minipage}
    \begin{minipage}{0.3\textwidth}
    \centering
    \begin{forest}
    for tree={
        grow=south,
        edge={-stealth},
        draw,
        rounded corners,
        node options={align=center},
        s sep=5mm,
        l sep=7mm
    }
    [2. geometry change, fill=lightgray
      [width change
      [left]
      [right]
      ]
      [border shape
      [left]
      [right]
      ]
    ]
    \end{forest}
    \end{minipage}
    \begin{minipage}{0.3\textwidth}
    \centering
    \begin{forest}
    for tree={
        grow=south,
        edge={-stealth},
        draw,
        rounded corners,
        node options={align=center},
        s sep=5mm,
        l sep=7mm
    }
    [3. lane type change, fill=lightgray
      [restriction
        [vehicle $\rightarrow$ ...]
      ]
      [opening
         [... $\rightarrow$ vehicle]
      ]
    ]
    \end{forest}
    \end{minipage}

    \vspace{0.5cm}
    \begin{minipage}{0.45\textwidth}
    \centering
    \begin{forest}
    for tree={
        grow=south,
        edge={-stealth},
        draw,
        rounded corners,
        node options={align=center},
        s sep=10mm,
        l sep=7mm
    }
    [4. insertion, fill=lightgray]
    \end{forest}
    \hspace{3cm}
    \begin{forest}
    for tree={
        grow=south,
        edge={-stealth},
        draw,
        rounded corners,
        node options={align=center},
        s sep=10mm,
        l sep=7mm
    }
    [5. deletion, fill=lightgray]
    \end{forest}
    \end{minipage}
    \caption{Hierarchical annotation granularity within atomic changes. Notably, insertions and deletions cannot be categorized further, as they do not have any correspondences between prior and ground-truth map.}\label{fig:trees}
\end{figure*}

While atomic changes remain indivisible by definition, we introduce additional levels of granularity beyond the five categories of atomic changes, to suffice desideratum C in \cref{tab:desiderata}. This hierarchy should be understood as a refinement in detail rather than a contradiction to their atomic nature.
Building upon the lane segment properties outlined in \cref{tab:lane_properties}, we structure hierarchical trees that further categorize atomic change categories (\cref{fig:trees}). These finer-grained annotations were not utilized in the main paper but are included as part of the dataset release to further enhance interpretability and usability.
\begin{table*}[h!]
    \centering
    \begin{tabular}{c|l|ccc|ccc|ccc}
        \toprule
        \multicolumn{2}{l|}{Category} & \multicolumn{3}{c|}{train} & \multicolumn{3}{c|}{val} & \multicolumn{3}{c}{test} \\
        \multicolumn{2}{l|}{} & global & frame & element & global & frame & element & global & frame & element \\
        \midrule
        \multicolumn{2}{l|}{total}     & 697  & 77670  & 1368387  & 102  & 11287  & 212695  & 111  & 10078  & 189331  \\
        \multicolumn{2}{l|}{of which changed}   & 697  & 43536  & 240849  & 102  & 5545  & 32555  & 104  & 5879  & 50459  \\
        \midrule
        \multirow{6}{*}{\rotatebox{90}{ls}} 
        & geometry  & 336  & 18027  & 68266  & 50  & 2043  & 7773  & 63  & 3491  & 21686  \\
        & mark      & 411  & 19789  & 42387  & 59  & 2359  & 5361  & 78  & 4243  & 14352  \\
        & insertion & 325  & 13842  & 49909  & 58  & 2318  & 8054  & 37  & 1862  & 5354  \\
        & deletion  & 212  & 8052   & 16171  & 37  & 1213  & 2419  & 27  & 1227  & 3121  \\
        & topology  & 149  & 5497   & 6767   & 25  & 635   & 829   & 36  & 1488  & 2527  \\
        & type      & 23   & 1539   & 4276   & 7   & 258   & 1035  & 6   & 416   & 1172  \\
        \midrule
        \multirow{3}{*}{\rotatebox{90}{pc}} 
        & geometry  & 169  & 5298   & 8591   & 25  & 823   & 1044  & 4   & 141   & 266   \\
        & insertion & 324  & 9945   & 17382  & 43  & 1258  & 2228  & 16  & 511   & 638   \\
        & deletion  & 483  & 15541  & 27100  & 80  & 2238  & 3812  & 20  & 790   & 1343  \\
        \bottomrule
    \end{tabular}
    \caption{Comparison of global, frame-wise, and element-wise change annotations across train, validation, and test sets. The change-class wise annotations are divided by lane segments (ls) and pedestrian crossings (pc).}
    \label{tab:train_val_test}
\end{table*}

\section{Dataset statistics}\label{sec:datasetstats}
In \cref{tab:train_val_test} we provide the details on annotations in ArgoTweak on global, frame and element-level. By global level, we mean the complete HD map for a specific driving sequence, whereas a frame indicates a $50\times50 \text{m}^2$ map patch around the local vehicle pose. We include the information for all atomic change categories, while in the experiments, type information has not been used. Topology denotes special geometric changes within intersections that maintain function on the road graph (see \cref{sec:topofunc} for details). 

\section{Rule-based prior generation}\label{sec:simpletbv}
\subsection{Pedestrian crossing perturbation procedure}
The implementation in our script largely follows the methodology described in \cite{tbv}, with some notable deviations. Below, we summarize the similarities:
\begin{itemize}
    \item \textbf{Sampling of lane segments:} The probability distribution is biased in favor of intersection lane segments, with a 4.5x higher weight, ensuring pedestrian crossings are more likely placed near intersections.
    \item \textbf{Waypoint sampling and orientation:} We interpolate the centerline of the sampled lane and select a random point, using the normal at that point to define the principal axis of the pedestrian crossing.
    \item \textbf{Road extent determination:} We compute the road polygon extent and identify the shortest valid span as \cite{tbv} suggests.
    \item \textbf{Width sampling:} A width value is sampled from a normal distribution $w \sim \mathcal{N}(3.5,1)$ and clipped to $[2,4]$ meters.
    \item \textbf{Overlap avoidance:} We ensure that no significant intersection with existing pedestrian crossings occurs, maintaining an IoU below 0.05.
    \item \textbf{Rendering of pedestrian crossings:} The pedestrian crossing is rendered as a buffered rectangle.
\end{itemize}

Despite the similarities, some deviations from the described methodology exist:
\begin{itemize}
    \item \textbf{Sampling iteration limit:} Instead of indefinitely sampling until success, our implementation enforces a maximum of 20 iterations for the global map, preventing infinite loops.
    \item \textbf{Width validation:} While the width is normally sampled and clipped, an additional height condition $(h > 2m)$ is enforced, which is not explicitly mentioned in the prior work.
    \item \textbf{Random crop constraints:} The original procedure ensures sampled waypoints avoid the outermost $\frac{1}{8}$ of the image; our script does not explicitly enforce this constraint, because this is not necessarily the case in real-world scenarios. Instead, we enforce that the pedestrian crossing intersects with a buffered area of 15 m around the ego vehicle's trajectory.
\end{itemize}

\subsection{Lane geometry perturbation procedure}
We implement lane modifications in accordance with the prior methodology, with some adaptations. Below, we outline the similarities and deviations.

\begin{itemize}
\item \textbf{Altering lane markings:} We modify lane boundary markings through transitions between solid and dashed, as well as between visible and non-visible markings.
\item \textbf{Modifying lane boundaries over multiple segments:} We iterate through connected lane segments, modifying three consecutive segments for marking changes and five for bike lanes.
\item \textbf{Adding bike lanes:} We identify the rightmost lane and divide it in half, introducing a new bike lane with solid white boundaries, as described in the original methodology.
\item \textbf{Deleting or modifying lane boundaries:} We selectively delete or alter painted lane boundary markings while ensuring that implicit boundaries remain unchanged.
\end{itemize}
We report the following deviations:
\begin{itemize}
\item \textbf{Ensuring perturbations stay within the field of view:} The original procedure enforces a constraint to avoid perturbations near the outermost $\frac{1}{8}$ of the rendered image, which we do not explicitly enforce. Instead, we again enforce that the modified segment intersects with a buffered area of 15 m around the ego vehicle's trajectory.
\item \textbf{Change frequency}: We insert a maximum number of 2 bike lanes per global map, and attempt to change four three-segment sequences of lane border markings.
\end{itemize}

\section{Additional results}\label{sec:moreresults}

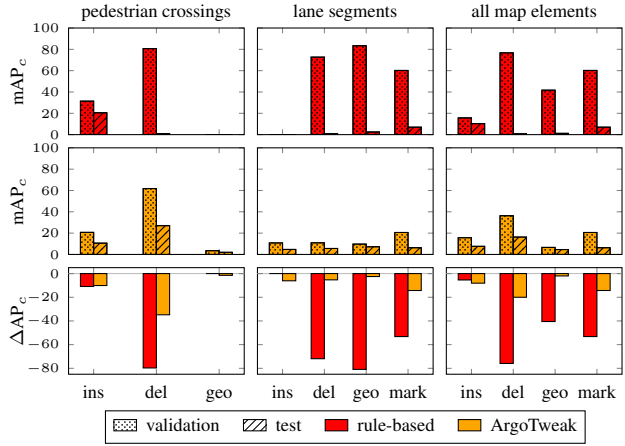
\begin{figure}  
    \centering
    \begin{tikzpicture}
        
       \begin{groupplot}[
    group style={group size=3 by 3, horizontal sep=5pt, vertical sep=5pt}, 
    ybar,
    symbolic x coords={ins,del,geo,mark},
    xtick=data,
    ymin=0, ymax=100,
    width=0.47\linewidth, 
    height=3cm,
    legend style={font=\tiny, draw=black, fill=white}, 
    enlarge x limits=0.2,
    trim axis left, trim axis right,
    tick style={inner sep=2pt, major tick length=2pt},
    tick align=inside,
    every axis x label/.style={yshift=-2pt, baseline} 
]

\nextgroupplot[
    title={pedestrian crossings},
    xlabel={}, xticklabels={}, 
    title style={font=\scriptsize, yshift=-7pt}, 
    ylabel={$\text{mAP}_c$}, 
    ylabel style={font=\scriptsize, yshift=-16pt}, 
    font=\tiny, bar width=5pt,
]
\addplot[fill=mplred, bar shift=-2.5pt, postaction={pattern=staggereddots, pattern color=black}] 
    coordinates {(ins,31.4) (del,80.7) (geo,0)};
\addplot[fill=mplred, bar shift=2.5pt, postaction={pattern=north east lines, pattern color=black}] 
    coordinates {(ins,20.5) (del,1) (geo,0)};

\nextgroupplot[title={lane segments},xlabel={}, xticklabels={},  title style={font=\scriptsize, yshift=-7pt}, ylabel={}, yticklabels={}, font=\scriptsize, bar width=5pt]
\addplot[fill=mplred, bar shift=-2.5pt, postaction={pattern=staggereddots, pattern color=black}] 
    coordinates {(ins,0.0) (del,72.7) (geo,83.4) (mark,60.2)};
\addplot[fill=mplred, bar shift=2.5pt, postaction={pattern=north east lines, pattern color=black}] 
    coordinates {(ins,0.0) (del,0.85) (geo,2.5) (mark,7.0)};

\nextgroupplot[title={all map elements}, xlabel={}, xticklabels={}, title style={font=\scriptsize, yshift=-7pt}, ylabel={}, yticklabels={}, font=\scriptsize, bar width=5pt]
\addplot[fill=mplred, bar shift=-2.5pt, postaction={pattern=staggereddots, pattern color=black}] 
    coordinates {(ins,15.7) (del,76.7) (geo,41.7) (mark,60.2)};
\addplot[fill=mplred, bar shift=2.5pt, postaction={pattern=north east lines, pattern color=black}] 
    coordinates {(ins,10.3) (del,0.79) (geo,1.2) (mark,7.0)};

\nextgroupplot[xlabel={}, xticklabels={}, ylabel={$\text{mAP}_c$}, 
    ylabel style={font=\scriptsize, yshift=-16pt}, font=\tiny, bar width=5pt]
\addplot[fill=mplblue, bar shift=-2.5pt, postaction={pattern=staggereddots, pattern color=black}] 
    coordinates {(ins,20.7) (del,61.8) (geo,3.5)};
\addplot[fill=mplblue, bar shift=2.5pt, postaction={pattern=north east lines, pattern color=black}] 
    coordinates {(ins,10.6) (del,27.0) (geo,2.0)};

\nextgroupplot[xlabel={}, xticklabels={}, ylabel={}, yticklabels={}, font=\scriptsize, bar width=5pt]
\addplot[fill=mplblue, bar shift=-2.5pt, postaction={pattern=staggereddots, pattern color=black}] 
    coordinates {(ins,10.8) (del,10.9) (geo,9.7) (mark,20.6)};
\addplot[fill=mplblue, bar shift=2.5pt, postaction={pattern=north east lines, pattern color=black}] 
    coordinates {(ins,4.7) (del,5.6) (geo,7.2) (mark,6.3)};

\nextgroupplot[xlabel={}, xticklabels={}, ylabel={}, yticklabels={}, font=\scriptsize, bar width=5pt]
\addplot[fill=mplblue, bar shift=-2.5pt, postaction={pattern=staggereddots, pattern color=black}] 
    coordinates {(ins,15.7) (del,36.3) (geo,6.6) (mark,20.6)};
\addplot[fill=mplblue, bar shift=2.5pt, postaction={pattern=north east lines, pattern color=black}] 
    coordinates {(ins,7.6) (del,16.3) (geo,4.6) (mark,6.3)};

\nextgroupplot[xticklabels={\strut ins, \strut del, \strut geo}, 
    ylabel={$\Delta\text{AP}_{c}$}, 
    ymin=-85, ymax=5,
    ylabel style={font=\scriptsize, yshift=-16pt}, 
    yticklabel style={font=\tiny}, xticklabel style={font=\scriptsize}, 
    font=\scriptsize, bar width=5pt, extra y ticks={0},
extra y tick labels={},
extra y tick style={grid=major, thin, black}]
\addplot[fill=mplred, bar shift=-2.5pt] 
    coordinates {(ins,-10.9) (del,-79.7) (geo,0.0)};
\addplot[fill=mplblue, bar shift=2.5pt] 
    coordinates {(ins,-10.1) (del,-34.8) (geo,-1.5)};

\nextgroupplot[ylabel={}, yticklabels={}, font=\scriptsize, bar width=5pt,  ymin=-85, ymax=5, xticklabels={\strut ins, \strut del, \strut geo, \strut mark},
extra y ticks={0},
extra y tick labels={},
extra y tick style={grid=major, thin, black}]
\addplot[fill=mplred, bar shift=-2.5pt] 
    coordinates {(ins,0.0) (del,-71.9) (geo,-80.9) (mark,-53.2)};
\addplot[fill=mplblue, bar shift=2.5pt] 
    coordinates {(ins,-6.1) (del,-5.3) (geo,-2.5) (mark,-14.3)};

\nextgroupplot[ylabel={}, yticklabels={}, font=\scriptsize, bar width=5pt,  ymin=-85, ymax=5, xticklabels={\strut ins, \strut del, \strut geo, \strut mark}, extra y ticks={0},
extra y tick labels={},
extra y tick style={grid=major, thin, black}]
\addplot[fill=mplred, bar shift=-2.5pt] 
    coordinates {(ins,-5.4) (del,-75.9) (geo,-40.5) (mark,-53.2)};
\addplot[fill=mplblue, bar shift=2.5pt] 
    coordinates {(ins,-8.1) (del,-20.0) (geo,-2) (mark,-14.3)};
  \end{groupplot}
\node[
    draw=black,
    fill=white,
    font=\scriptsize, 
    inner sep=3pt, 
    align=center, 
    yshift=-20pt, 
] at ($(group c1r3.south)!0.5!(group c3r3.south)$) 
{ \tikz \draw[fill=white, draw=black, pattern=staggereddots, pattern color=black] (0,0) rectangle (0.3,0.15); \ validation \quad
    \tikz \draw[fill=mplred!20, draw=black, pattern=north east lines, pattern color=black] (0,0) rectangle (0.3,0.15); \ test
    \quad
    \tikz \draw[fill=mplred] (0,0) rectangle (0.3,0.15); \ rule-based
    \quad
    \tikz \draw[fill=mplblue] (0,0) rectangle (0.3,0.15); \ ArgoTweak};
    \end{tikzpicture}
\caption{Sim2real gap computed on $\text{mAP}_c$.}\label{fig:sim2real_AP}
\end{figure}
In \cref{tab:filtered_metrics_split}, we provide the evaluation of our model on the test split of ArgoTweak with primary and secondary change assessment heads active. This complements \cref{tab:ablation_tab} in the main paper while providing the $\text{AP}_c$ values for lane segments and pedestrian crossings objects separately.

In \cref{fig:sim2real_AP}, we present an analysis of the sim-to-real gap based on $\text{mAP}_c$. While the reduction in the sim-to-real gap is clearly observable in this figure, the effects of map generation and change detection are intertwined, making interpretation more complex. The task is inherently more challenging for $\text{mAP}_c$ than for $\text{mAcc}_c$, as the model must accurately assess the change status while simultaneously capturing the correct geometric representation. For $\text{mAP}_c$ calculation, we use all elements, while for $\text{mAcc}_c$, we threshold at a confidence score of $0.5$. 

\begin{figure*}
    \centering
    \setlength{\tabcolsep}{0pt}
    \begin{minipage}{0.45\textwidth}
        \begin{subfigure}{1\textwidth}
        \centering
        \fboxsep=0pt
        {{\includegraphics[width=\textwidth]{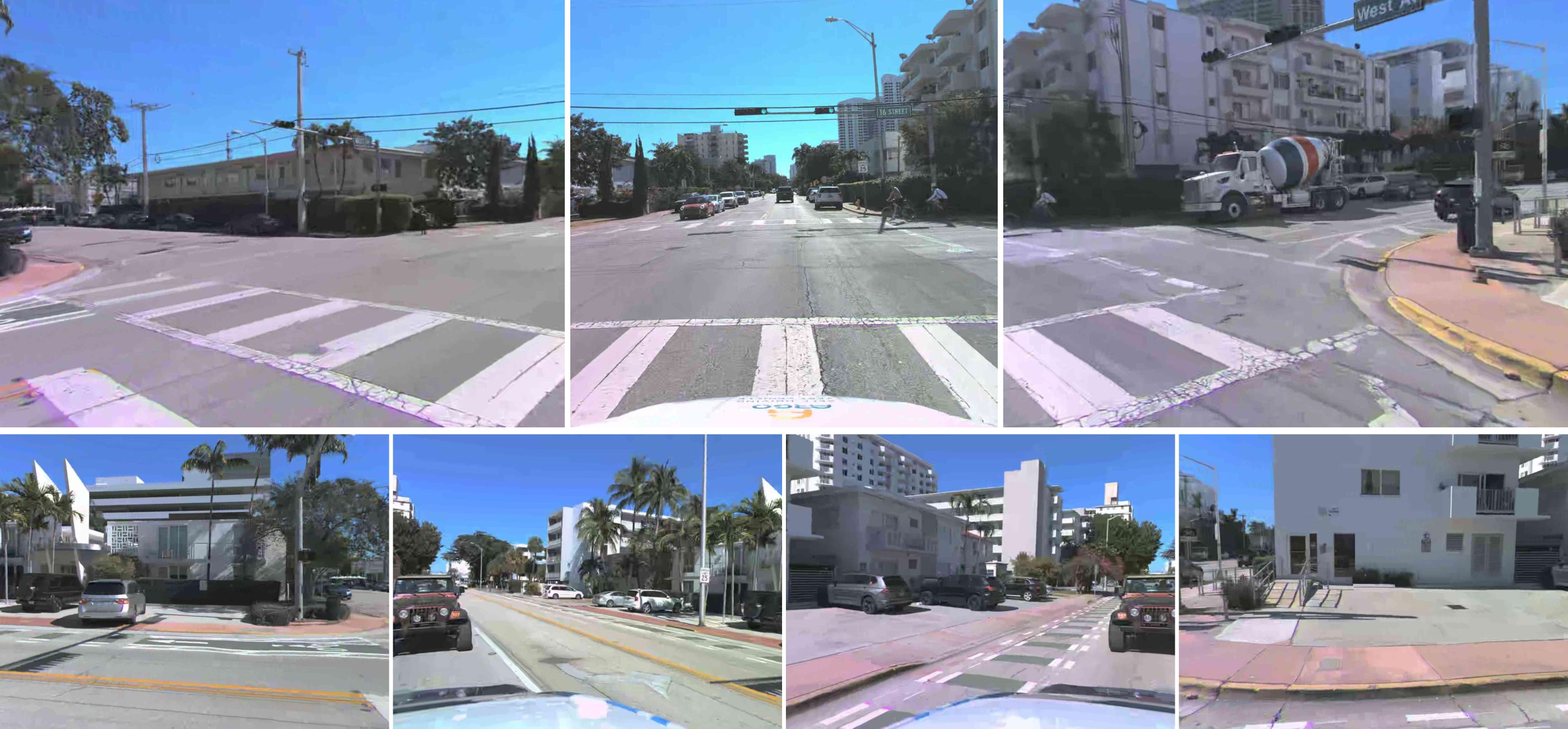}}}
        \caption{sensor input}
        
    \end{subfigure}
    \end{minipage}
    \hfill
    \begin{minipage}{0.16\textwidth}
        \begin{subfigure}{1\textwidth}
        \centering
        \fboxsep=0pt
        {\rotatebox{-90}{\includegraphics[width=\textwidth]{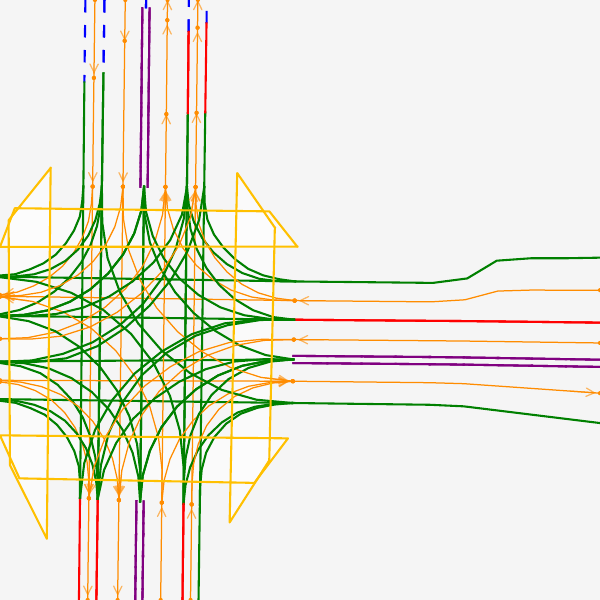}}}
        \caption{map prior}
        
    \end{subfigure}
    \end{minipage}
    \hfill
    \begin{minipage}{0.34\textwidth}
        \centering
        \begin{minipage}{0.48\textwidth}
            \begin{subfigure}{1\textwidth}
        \centering
        \fboxsep=0pt
        {\rotatebox{-90}{\includegraphics[width=\textwidth]{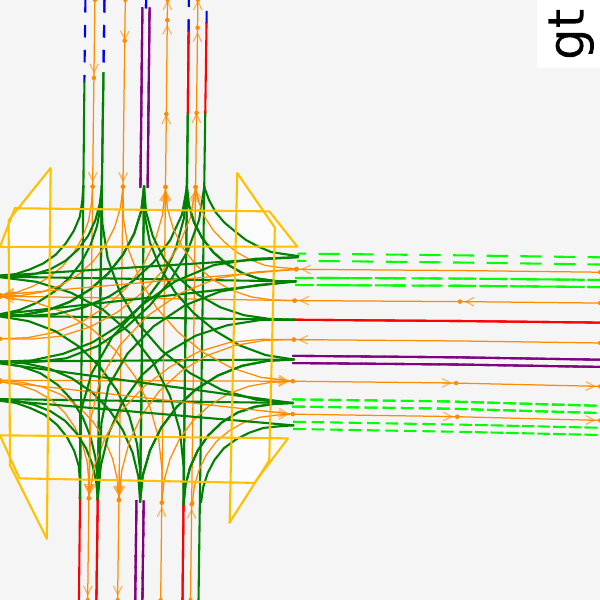}}}
        
    \end{subfigure}
        \end{minipage}
        \hspace{-0.3\baselineskip}
        \begin{minipage}{0.48\textwidth}
            \centering
            \rotatebox{-90}{\adjustbox{padding=0.5pt}{\includegraphics[width=\textwidth]{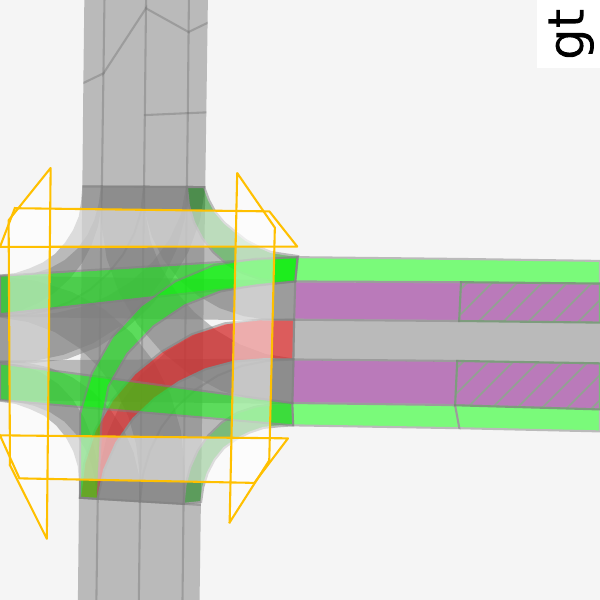}}}
        \end{minipage}
        
        \vspace{-0.1\baselineskip}
        
        \begin{minipage}{0.48\textwidth}
             \begin{subfigure}{1\textwidth}
        \centering
        \fboxsep=0pt
        {\rotatebox{-90}{\includegraphics[width=\textwidth]{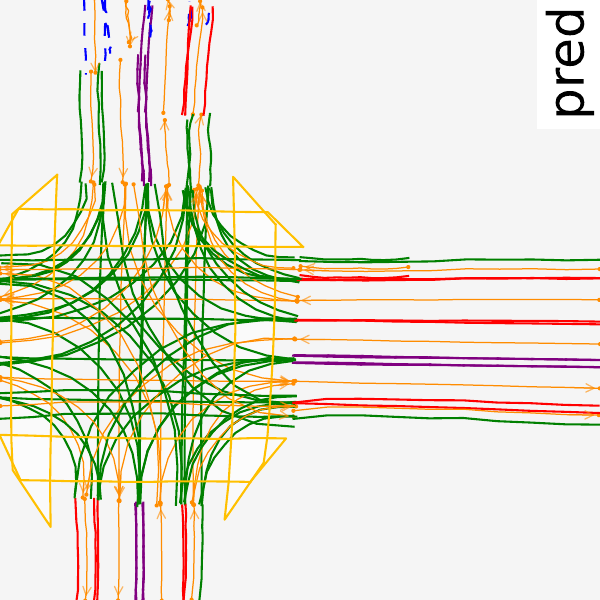}}}
        \caption{map update}
        
    \end{subfigure}
        \end{minipage}
        \hspace{-0.3\baselineskip}
        \begin{minipage}{0.48\textwidth}
            \begin{subfigure}{1\textwidth}
        \centering
        \fboxsep=0pt
        {\rotatebox{-90}{\includegraphics[width=\textwidth]{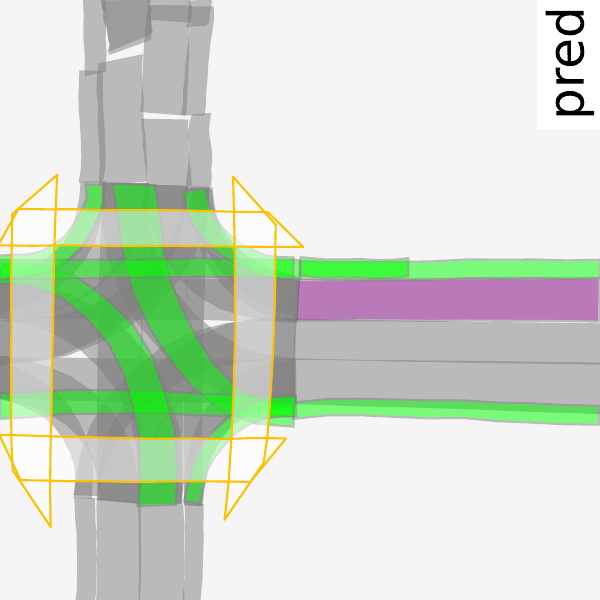}}}
        
            \caption{change assessment}
            \end{subfigure}
        \end{minipage}
    \end{minipage}

     \vspace{\baselineskip}

    \begin{minipage}{0.45\textwidth}
        \begin{subfigure}{1\textwidth}
        \centering
        \fboxsep=0pt
        {{\includegraphics[width=\textwidth]{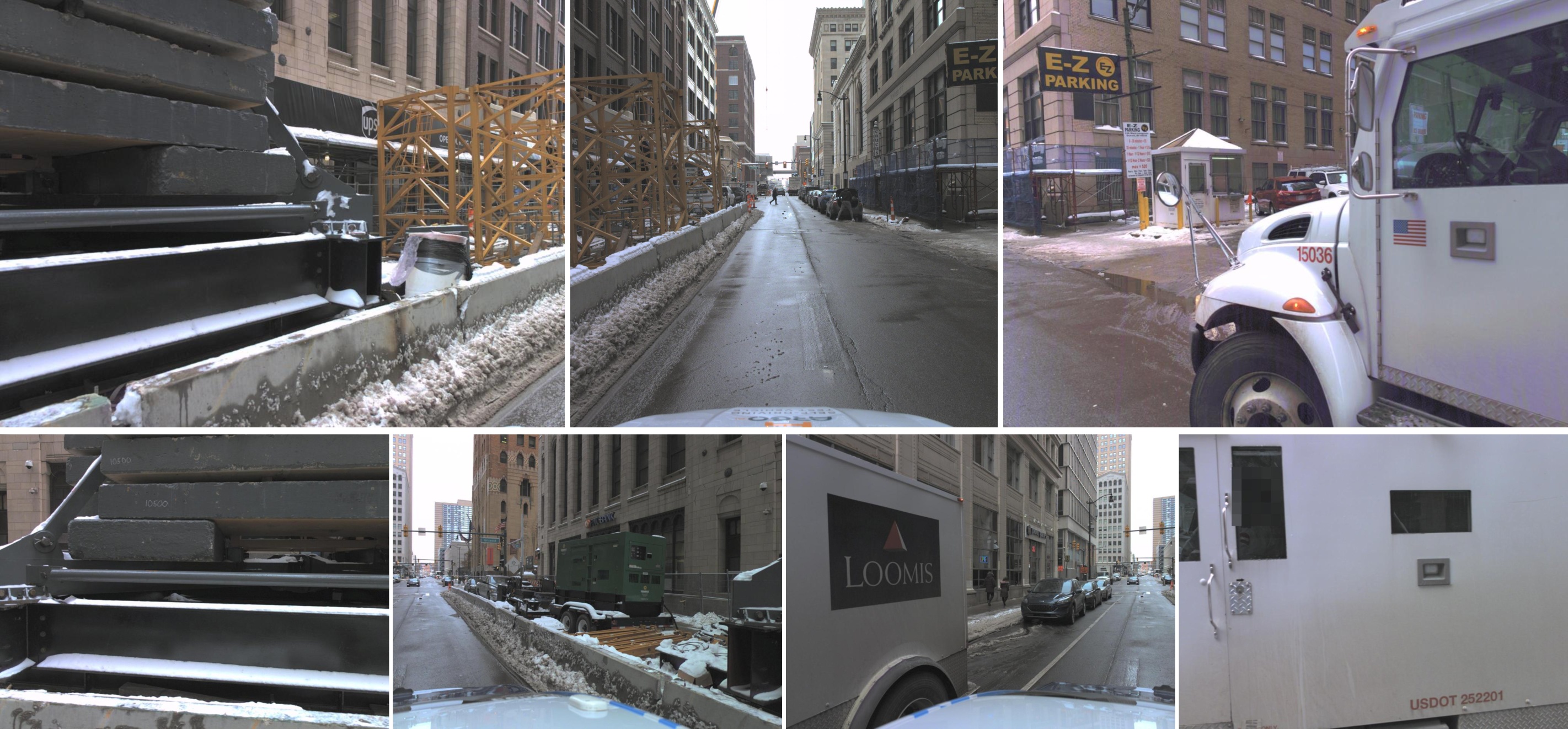}}}
        \caption{sensor input}
        
    \end{subfigure}
    \end{minipage}
    \hfill
    \begin{minipage}{0.16\textwidth}
        \begin{subfigure}{1\textwidth}
        \centering
        \fboxsep=0pt
        {\rotatebox{-90}{\includegraphics[width=\textwidth]{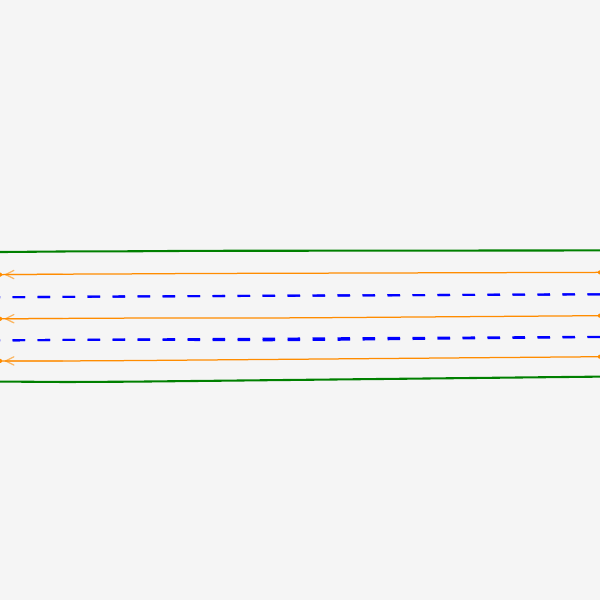}}}
        \caption{map prior}
        
    \end{subfigure}
    \end{minipage}
    \hfill
    \begin{minipage}{0.34\textwidth}
        \centering
        \begin{minipage}{0.48\textwidth}
            \begin{subfigure}{1\textwidth}
        \centering
        \fboxsep=0pt
        {\rotatebox{-90}{\includegraphics[width=\textwidth]{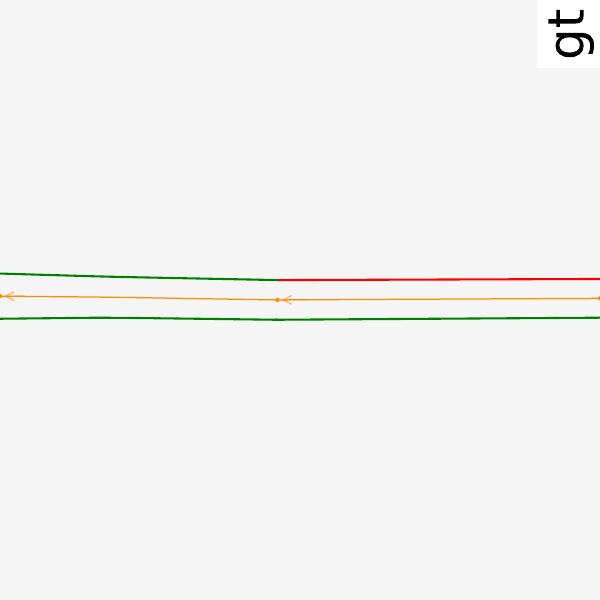}}}
        
    \end{subfigure}
        \end{minipage}
        \hspace{-0.3\baselineskip}
        \begin{minipage}{0.48\textwidth}
            \centering
            \rotatebox{-90}{\adjustbox{padding=0.5pt}{\includegraphics[width=\textwidth]{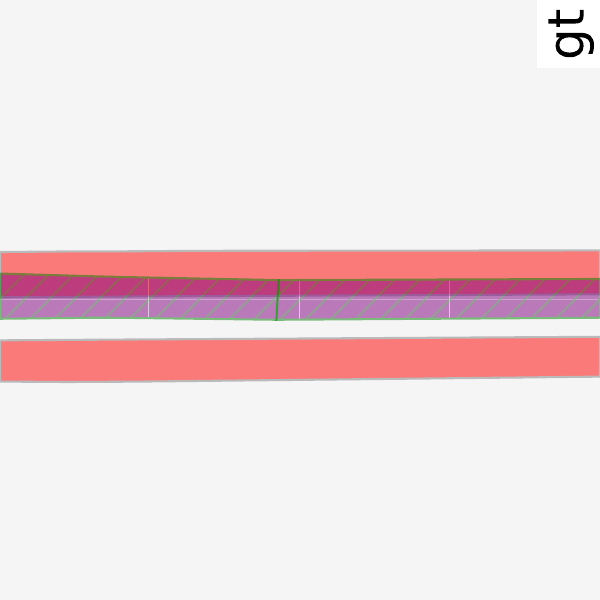}}}
        \end{minipage}
        
        \vspace{-0.1\baselineskip} 
        
        \begin{minipage}{0.48\textwidth}
             \begin{subfigure}{1\textwidth}
        \centering
        \fboxsep=0pt
        {\rotatebox{-90}{\includegraphics[width=\textwidth]{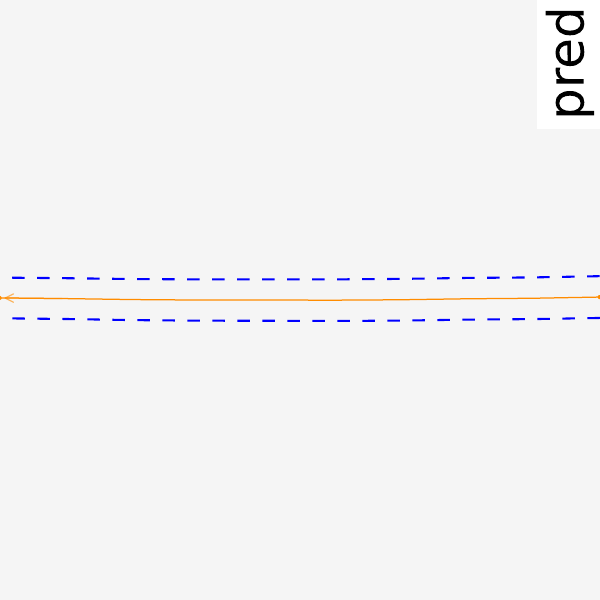}}}
        \caption{map update}
        
    \end{subfigure}
        \end{minipage}
        \hspace{-0.3\baselineskip}
        \begin{minipage}{0.48\textwidth}
            \begin{subfigure}{1\textwidth}
        \centering
        \fboxsep=0pt
        {\rotatebox{-90}{\includegraphics[width=\textwidth]{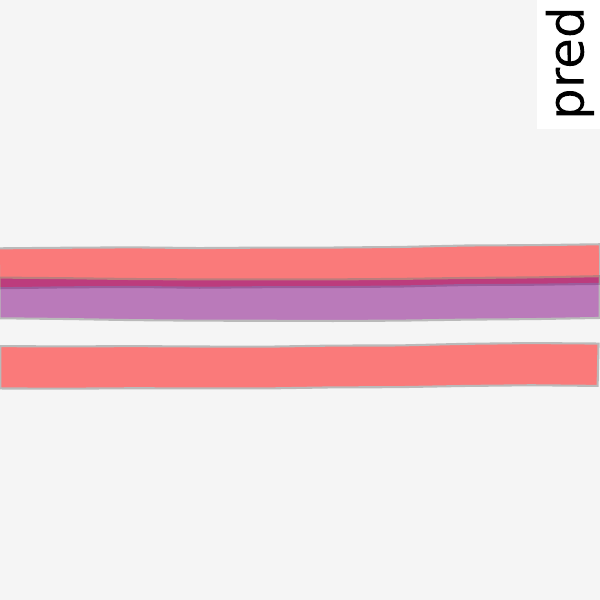}}}
        
            \caption{change assessment}
            \end{subfigure}
        \end{minipage}
    \end{minipage}

     \vspace{\baselineskip}

    \begin{minipage}{0.45\textwidth}
        \begin{subfigure}{1\textwidth}
        \centering
        \fboxsep=0pt
        {{\includegraphics[width=\textwidth]{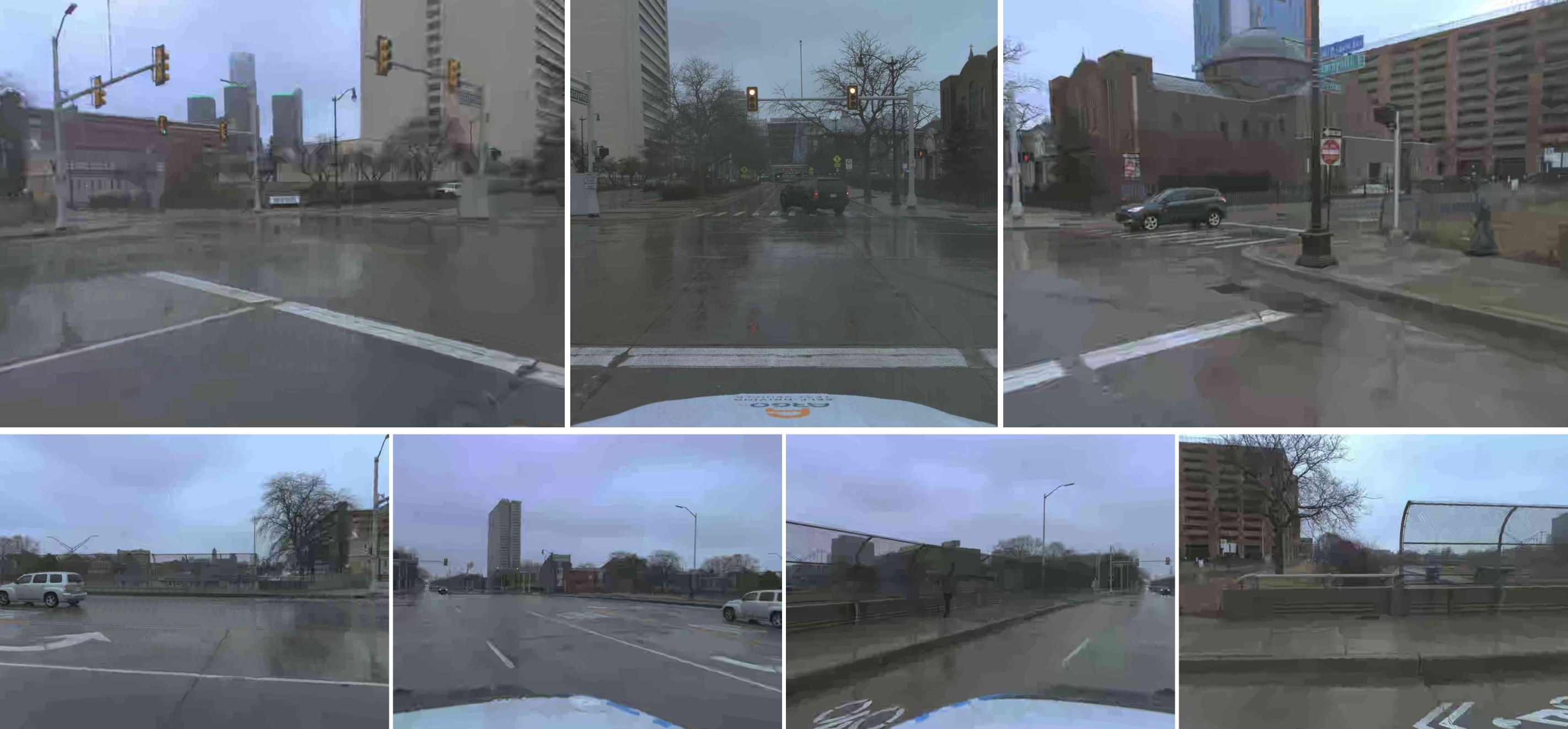}}}
        \caption{sensor input}
        
    \end{subfigure}
    \end{minipage}
    \hfill
    \begin{minipage}{0.16\textwidth}
        \begin{subfigure}{1\textwidth}
        \centering
        \fboxsep=0pt
        {\rotatebox{-90}{\includegraphics[width=\textwidth]{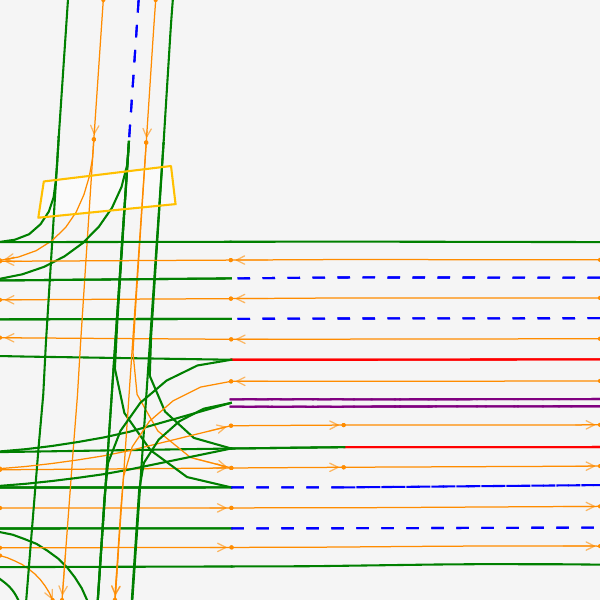}}}
        \caption{map prior}
        
    \end{subfigure}
    \end{minipage}
    \hfill
    \begin{minipage}{0.34\textwidth}
        \centering
        \begin{minipage}{0.48\textwidth}
            \begin{subfigure}{1\textwidth}
        \centering
        \fboxsep=0pt
        {\rotatebox{-90}{\includegraphics[width=\textwidth]{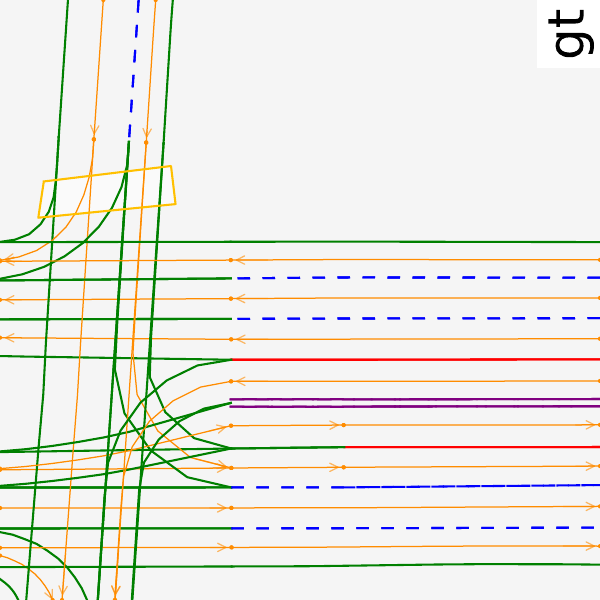}}}
        
    \end{subfigure}
        \end{minipage}
        \hspace{-0.3\baselineskip}
        \begin{minipage}{0.48\textwidth}
            \centering
            \rotatebox{-90}{\adjustbox{padding=0.5pt}{\includegraphics[width=\textwidth]{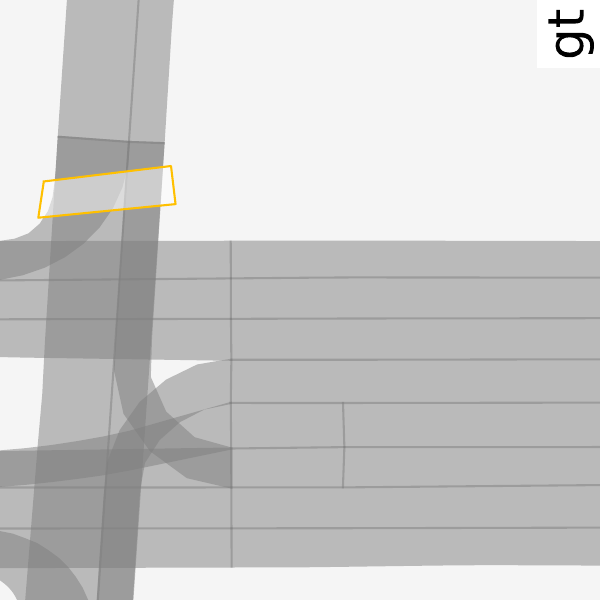}}}
        \end{minipage}
        
        \vspace{-0.1\baselineskip} 
        
        \begin{minipage}{0.48\textwidth}
             \begin{subfigure}{1\textwidth}
        \centering
        \fboxsep=0pt
        {\rotatebox{-90}{\includegraphics[width=\textwidth]{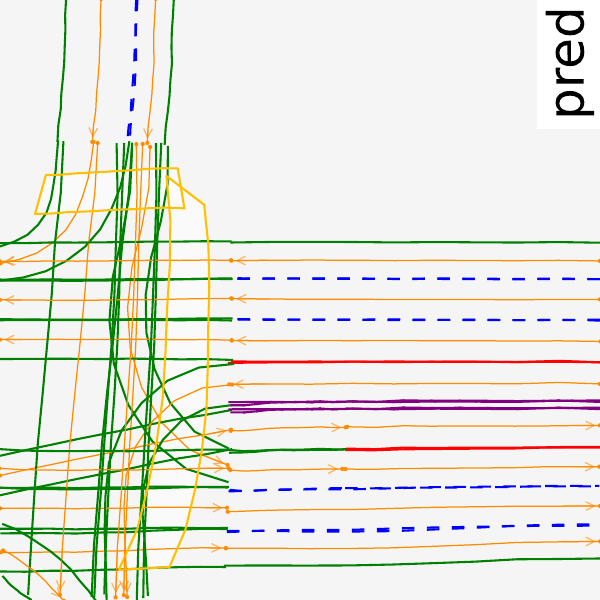}}}
        \caption{map update}
        
    \end{subfigure}
        \end{minipage}
        \hspace{-0.3\baselineskip}
        \begin{minipage}{0.48\textwidth}
            \begin{subfigure}{1\textwidth}
        \centering
        \fboxsep=0pt
        {\rotatebox{-90}{\includegraphics[width=\textwidth]{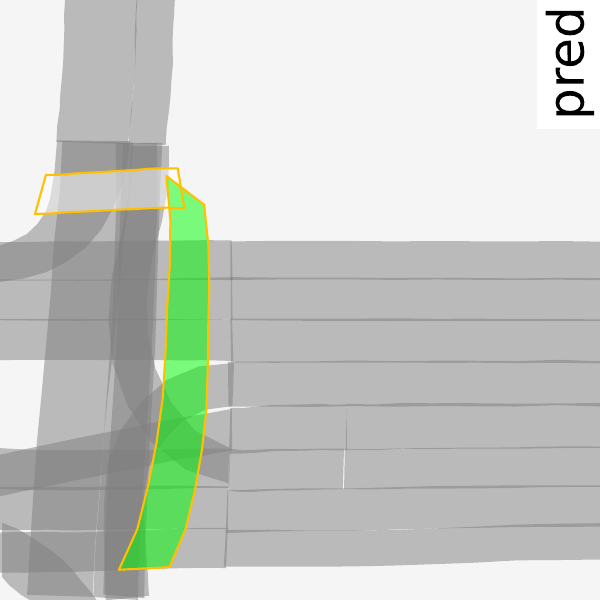}}}
        
            \caption{change assessment}
            \end{subfigure}
        \end{minipage}
    \end{minipage}

    \caption{Qualitative examples of our ArgoTweak-trained model. For change assessment, purple denotes lane marking changes, light green insertions, red deletions, dark green geometry changes. Striped elements indicate multiple changes per segment. In the first example, the model correctly detects the newly added bike lanes. In the second example, the road shape is correctly updated, but the appearance is not. In the third example, we observe how the model correctly classifies all lane segments as unchanged and reproduces them with high accuracy. However, the white stopline in front of the vehicle is mistaken for a pedestrian crossing insertion.}\label{fig:example2}
\end{figure*}

In \cref{fig:example2}, we provide more qualitative examples.
\begin{table}[t]
    \centering
    \begin{tabular}{l|cc|c}
        \toprule
        all categories &  $\text{AP}_c^\text{ls}$ & $\text{AP}_c^\text{pc}$ & mAP \\
        \midrule
        total & 75.4 & 82.2 & 78.8 \\
        \midrule
        change & 9.0 & 19.0 & 14.0 \\
        no change & 74.5 & 82.5 & 78.5 \\
        \midrule
        insertion & 4.7 & 10.6 & 7.6 \\
        deletion & 5.6 & 27.0 & 16.3 \\
        geometry & 7.2 & 2.0 & 4.6\\
        mark & 6.3 & -- & 6.3 \\
        \bottomrule
    \end{tabular}
    \caption{Performance of our model trained on ArgoTweak and evaluated on the ArgoTweak test set.}
    \label{tab:filtered_metrics_split}
\end{table}
\newpage

\end{document}